\documentclass{article} % For LaTeX2e
\usepackage{iclr2026_conference,times}
\usepackage{pifont}
\usepackage{amsmath,amsfonts,bm}

\def\eqref#1{equation~\ref{#1}}
\def\1{\bm{1}}

\DeclareMathAlphabet{\mathsfit}{\encodingdefault}{\sfdefault}{m}{sl}
\SetMathAlphabet{\mathsfit}{bold}{\encodingdefault}{\sfdefault}{bx}{n}

\usepackage{enumitem}
\usepackage{hyperref}
\usepackage{url}
\usepackage{graphicx}
\usepackage{booktabs}
\usepackage{amsmath}
\usepackage{amssymb}
\usepackage{xspace}
\usepackage{xcolor}
\usepackage{caption}
\usepackage{subcaption}
\usepackage{amsthm}
\usepackage{multirow}
\usepackage{wrapfig}
\usepackage{listings}
\lstdefinestyle{promptstyle}{
  basicstyle=\ttfamily\scriptsize,
  breaklines=true,
  breakautoindent=false,
  columns=fullflexible,
  keepspaces=true,
  frame=single,
  rulecolor=\color{gray!55},
  backgroundcolor=\color{gray!8},
  framexleftmargin=3pt,
  framexrightmargin=3pt,
  xleftmargin=4pt,
  xrightmargin=4pt,
  aboveskip=7pt,
  belowskip=7pt,
}
\title{
\raisebox{-0.25\height}{\includegraphics[height=1.5em]{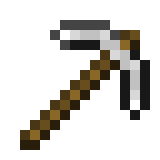}}
\
VibeWorlding: Can Multimodal Agents \\ Construct 3D Open Worlds End-to-End?}

\author{%
  Yansong Ning\textsuperscript{1}\thanks{Work done during internship at Tencent.},
Jingwen Ye\textsuperscript{2},
Zhongkai Wu\textsuperscript{2},
Yang Sun\textsuperscript{2},
  Yiqin Zhu\textsuperscript{2},
  Xingyi Li\textsuperscript{2} \\
  \textbf{
  Weidong Zhang\textsuperscript{2},
  Hao Liu\textsuperscript{1}\thanks{Corresponding author.}}
  \\
  \textsuperscript{1} AI Thrust, HKUST(GZ),
  \textsuperscript{2} TEG AIPD, Tencent \\
  \texttt{yning092connect.hkust-gz.edu.cn}, \texttt{liuh@ust.hk}\\ \texttt{\{jingwenye,wadewdzhang\}@tencent.com}
}

\newcommand{\method}{\textsc{VibeWorlding}\xspace}
\newcommand{\bench}{\textsc{VWE-Bench}\xspace}
\newcommand{\gym}{\textsc{VibeWorlding-Gym}\xspace}
\newtheorem{myDef}{Definition}
\newtheorem{pro}{Problem}

\iclrfinalcopy % Uncomment for camera-ready version, but NOT for submission.
\begin{document}

\maketitle
\vspace{-25pt}
\begin{center}
\small
\begin{tabular}{@{}r@{~}l@{}}
\raisebox{-0.25\height}{\includegraphics[height=1.05em]{Figure/ours.png}} & \textbf{Project Page:} \url{https://usail-hkust.github.io/VibeWorlding-Gym/} \\
[2pt]
\raisebox{-0.25\height}{\includegraphics[height=1.05em]{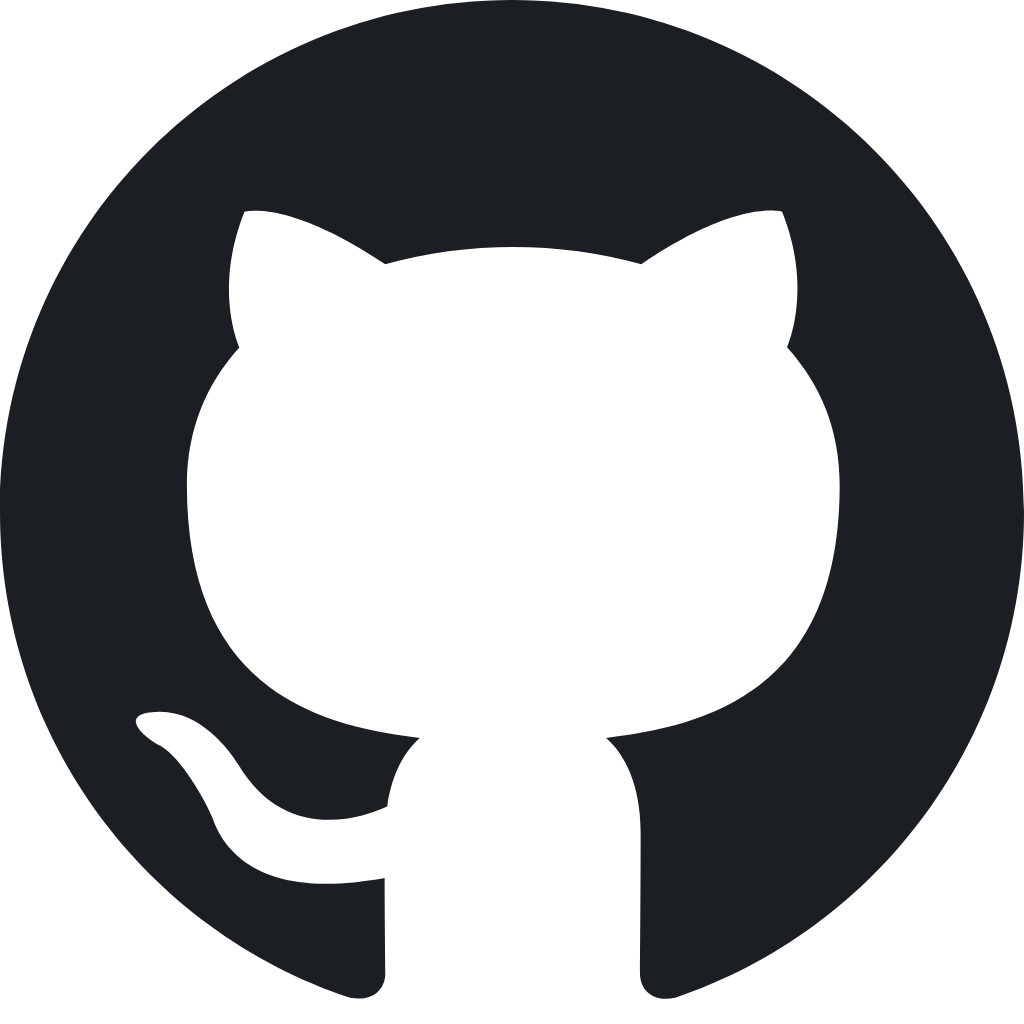}} & \textbf{Code:} \url{https://github.com/usail-hkust/VibeWorlding-Gym} \\[2pt]
\raisebox{-0.25\height}{\includegraphics[height=1.05em]{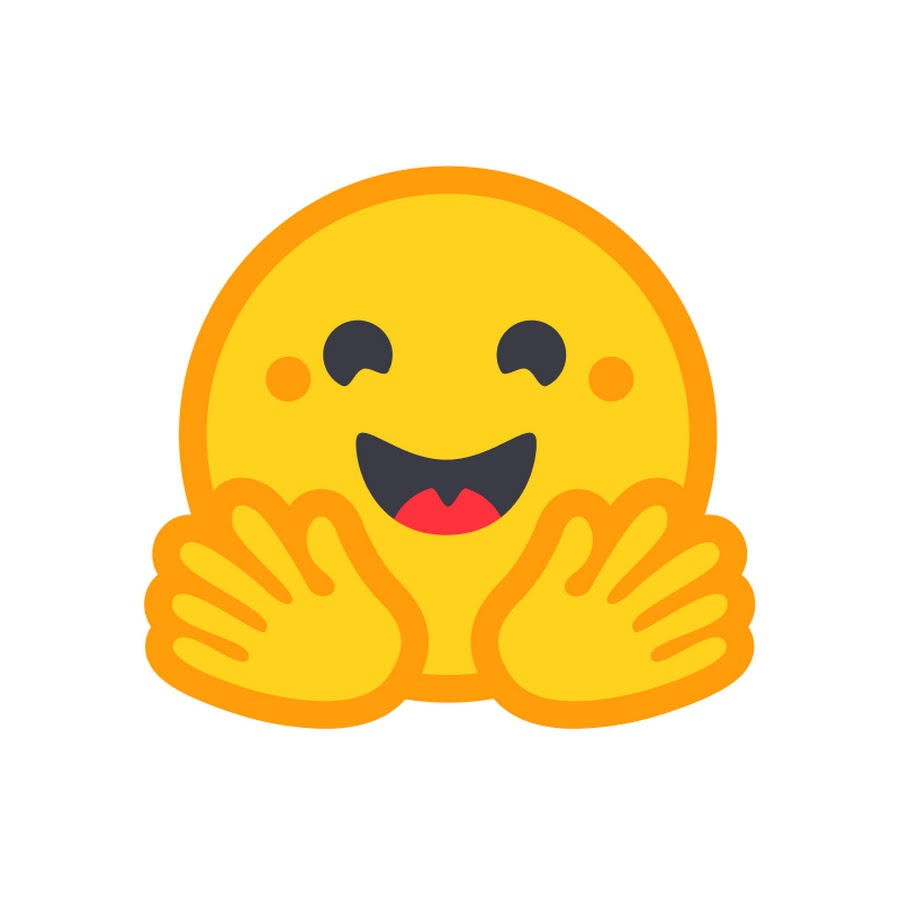}} & \textbf{Model:} \url{https://huggingface.co/collections/usail-hkust/vibeworlder} \\[2pt]
\raisebox{-0.25\height}{\includegraphics[height=1.05em]{Figure/huggingface.png}} & \textbf{Dataset:} \url{https://huggingface.co/datasets/usail-hkust/VWE-Bench} \\
\end{tabular}
\end{center}

\begin{abstract}
Constructing an interactive 3D open world from a user query is important for gaming, simulation, and embodied AI. 
However, existing methods are primarily evaluated on idealized, simple queries, making it difficult to systematically analyze and compare how multimodal agents understand user intent, use 3D tools, and reason over textual and visual 3D world information.
Furthermore, the absence of an open-source framework also hinders the systematic study of whether training (e.g., agentic RL post-training) can improve these underlying capabilities.
To this end, we propose \textbf{\method}, a unified framework for benchmarking and training vibe worlding agents: a multimodal agent that can autonomously infer user intent, plan scene layout, invoke 3D tools (e.g., asset retrieval/edit), and reflect on the multimodal feedback (e.g., 3D map and rendered 3D world images) in a multi-turn agent-environment interaction process.
To achieve this, we first build \textbf{\bench} (\textbf{V}ibe\textbf{W}orlding \textbf{E}valuation Benchmark), a benchmark of \underline{2,616} high-quality 3D assets, \underline{323} human-annotated seed 3D worlds, and \underline{6,828} reverse-synthesized multimodal user queries, split into verified queries with ground-truth and unverified queries with carefully designed rubrics.
Moreover, we develop \textbf{\gym}, a joint multimodal RL post-training framework that integrates (1) a sandbox environment unifying asset retrieval, editing, and image rendering as MCP tools, and (2) a rubric-based verifier that combines physical feasibility (e.g., asset collision detection) and intent fulfillment verification (e.g., user intent), supporting both fair model evaluation and scalable multimodal RL reward service.
Our experiments show that current frontier MLLMs are far from solving the vibe worlding agent task, with even GPT-5.5 and Qwen3.8-Max reaching below 60\% success rate, and trace the bottleneck to precise 3D world editing.
We further find that RL training can ease this weakness and enable open-source MLLMs to even surpass closed-source frontiers: our VibeWorlder-8B is comparable to frontier MLLMs, while our flagship VibeWorlder-30B-A3B attains the best overall Pass@1 among all evaluated models.
We release our data, code, and models to facilitate research for end-to-end 3D world construction.
\end{abstract}

\section{Introduction}
\label{sec:intro}
Constructing interactive 3D open worlds~\citep{wen2025survey} using predefined 3D assets is of great importance for gaming, simulation, and embodied
AI application~\citep{wu2026production}. 
Recently, with the rise of multimodal large
language model (MLLM), a growing body of work builds multimodal agents for automating 3D world construction from user queries. 
MLLM-powered 3D world construction agents have attracted broad interest across both academia and industry.

Existing MLLM-powered 3D world construction works fall into two categories.
The first decomposes the construction process into a sequential pipeline and
assigns a specialized sub-agent to each stage. 
For example, SceneCraft~\citep{hu2024scenecraft} and 3D-GPT~\citep{sun2024gpt3d} first employ a planner agent to generate the scene layout, and then use two additional agents to iteratively critique and refine the resulting 3D scene.
The second adopts an agentic workflow that enables an agent to autonomously plan, invoke
asset operation tools, and self-refine the 3D world over multi-turn interaction.
For example, SAGE~\citep{xia2026sage}, SceneWeaver~\citep{yang2025sceneweaver} and SceneReVis~\citep{zhao2026scenerevis}
integrate asset editing, procedural content generation, and rendering into a tool
set, and prompt the agent to interact with these tools for 3D world construction.
However, these works predominantly handle idealized, simple queries and \emph{struggle to support complex user queries with open-ended intents} in real-world scenarios.
In addition, \emph{most existing frameworks remain closed-source}, and there is still no unified, open-source framework for systematically benchmarking and training 3D world construction agents.

Inspired by recent advances such as Kimi K3 \citep{kimi2026k3}, which demonstrates strong 3D reasoning/vision capabilities for end-to-end 3D open world construction, we aim to develop a unified framework for benchmarking and training such agents, which we term \textbf{vibe worlding agents}.
However, it is non-trivial due to two factors. 
\emph{(1) The 3D asset environment is fragmented:} High-quality, physically consistent 3D assets are hard to obtain at scale.
Furthermore, the asset retrieval, editing, and rendering operations required by agents are scattered across incompatible tools, hindering end-to-end construction.
There are no existing works that unify them under one sandbox interface. 
\emph{(2) 3D world construction is inherently challenging to verify:} A feasible 3D world should adhere to physical constraints, such as realistic asset heights and collision-free layouts, while simultaneously satisfying user intent, task requirements, and aesthetic coherence. 
Such multifaceted criteria are difficult to evaluate using either handcrafted rules or a generic LLM judge alone.

\begin{figure}
\vspace{-25pt}
    \centering
    \includegraphics[width=1\linewidth]{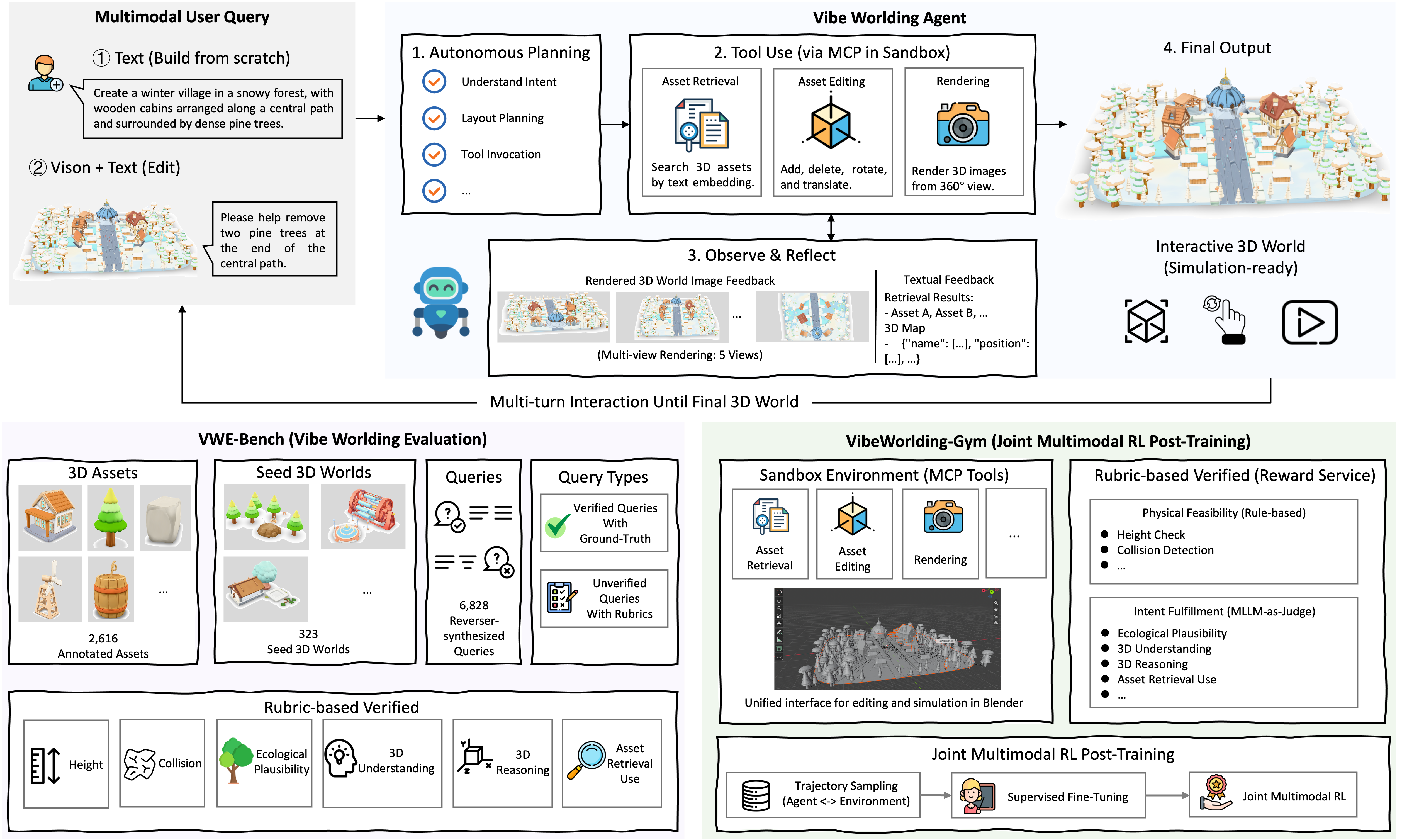}
    \caption{The overview of VWE-Bench and VibeWorlding-Gym.}
    \label{fig_intro}
\vspace{-15pt}
\end{figure}

To address these challenges, we present \textbf{\method}, a unified framework for benchmarking and training vibe worlding agents shown in Figure \ref{fig_intro}. 
As can be seen, we formulate the vibe worlding task as a multi-turn, multimodal, tool-integrated reasoning process: given a multimodal user query, either plain text to build a 3D world from scratch or an existing 3D world plus a textual instruction to edit it, the agent should autonomously infer user intent, plan the scene layout, invoke 3D tools (e.g., asset retrieval and editing), and reflect on the multimodal feedback (e.g., 3D map and rendered 3D world images) over multiple turns, until it emits a final interactive 3D world.

Specifically, we first build \textbf{\bench} (VibeWorlding Evaluation Benchmark), a
benchmark of 2,616 high-quality 3D assets, 323 human-annotated seed 3D worlds, and
6,828 reverse-synthesized user queries, split into verified queries with
ground-truth and unverified queries with carefully designed rubrics.
Through a dual-constraint verifier spanning both of physical feasibility check (i.e., height, collision) and intent fulfillment verification (i.e., ecological plausibility, 3D understanding, 3D reasoning, and asset retrieval), the VibeWorlding task can be systematically evaluated.
Furthermore, we develop \textbf{\gym}, a joint multimodal RL training framework that integrates a sandbox environment, which unifies asset retrieval, editing, and image rendering as MCP tools and the aforementioned rubric-based verifier. 
Built on the sandbox and verifier, we provide a unified pipeline for trajectory sampling, SFT, and multimodal RL, so that model evaluation and training are driven by the unified environment and reward signal.

We conduct a comprehensive experiment on \method to validate how existing multimodal agents perform on vibe worlding tasks. 
Our results reveal that: first, current models are far from solving the task: even GPT-5.5 and Qwen3.8-Max reach below $60\%$ on \bench.
Second, our six-capability analysis traces this gap to a bottleneck in 3D reasoning, where the model can understand the 3D world and user intention, but fails in accurately editing the world through 3D tools. 
Third, when reliable reward signals are available, agentic RL post-training can mitigate this weakness, enabling open MLLMs to match or even outperform frontier closed-source models. For example, it allows a small-sized model (our VibeWorlder-8B, post-trained from Qwen3-VL-8B) to reach parity with Gemini~3.1-pro, and our flagship VibeWorlder-30B-A3B to attain a better overall Pass@1 score than frontier MLLMs (e.g., GPT-5.5 and Qwen3.8-Max).

In summary, our contributions are:
\begin{itemize} [leftmargin=*, nosep]
\item \textbf{We release the first agentic RL framework for end-to-end 3D world construction tasks}, providing a unified stack: 2,616 annotated 3D assets, 323 seed 3D worlds, 6,828 user queries, an asset retrieval embedding model, an interactive sandbox environment, a dual-constraint verifier, and a unified agent post-training framework (including both SFT and agentic RL training).

\item \textbf{A comprehensive six-capability analysis.} 
We systematically measure multimodal agents on end-to-end 3D world construction tasks along physical feasibility and user intention fulfillment verification, to reveal their strengths and defects.
And we further analyze which of these capabilities multimodal RL post-training actually improves.

\item \textbf{The state-of-the-art results.} 
Our RL post-training enables open MLLMs to match and surpass industry closed-source frontier models on \bench: VibeWorlder-8B (post-trained from Qwen3-VL-8B) matches Gemini~3.1-pro, while VibeWorlder-30B-A3B achieves the best overall Pass@1 among all evaluated models, edging out GPT-5.5 and Qwen3.8-Max.
\end{itemize}

% We release our repository to advance the frontier of 3D generative intelligence.

\section{Preliminary}
\label{sec:prelim}

\subsection{Definition}
\label{sec:def}

We begin with the definition of 3D asset, 3D world, multimodal query, and 3D tools, and summarize them with illustrative examples in Table~\ref{tab:notation}.

\begin{myDef}
\textbf{3D asset.}
A 3D asset $m$ is the atomic building block of a 3D world and is defined as a
5-tuple $m =(\texttt{id}, \texttt{name}, \texttt{category}, \texttt{face\_count}, \texttt{bbox})$, where \texttt{id} is a unique asset id, \texttt{face\_count} is its mesh complexity, and \texttt{bbox} is the native bounding box that specifies its physical size. 
For example, \texttt{["00001", "Bookshelf\_01", "Furniture", "388", ([0.69, 0, 0], [48.22, 140.88, 131.42])]} denotes a bookshelf asset whose face count is 388, category is Furniture, and physical footprint is defined by the bounding box spanning from \texttt{(0.69, 0, 0)} to \texttt{(48.22, 140.88, 131.42)}.
\end{myDef}

\begin{myDef}
\textbf{3D world.} 
A 3D world $W$ is a set of placed assets and is defined as $W = \{(m, \texttt{pos[x,y,z]}, \texttt{rot[x,y,z]}, \texttt{sca}), \dots\}$, where each asset m is instantiated at the spatial location \texttt{pos[x,y,z]}, rotated by Euler angles \texttt{rot[x,y,z]} around the x-, y-, and z-axes, respectively, and scaled by a factor \texttt{sca}.
In this paper, a 3D world can be expressed in two complementary modalities: a textual
map $W_{\text{text}}$ that records the detailed location of each asset, and a set of images $W_{\text{image}}$ rendered from five viewpoints (front, back, left, right, and top-down). 
\end{myDef}

\begin{myDef}
\textbf{Multimodal query.} 
A multimodal query $q$ is defined in one of two forms: a plain text that prompts the agent to build a 3D world from scratch, or an existing 3D world paired with a textual instruction
$q=(W_{\text{text}}, W_{\text{image}}, \texttt{textual instruction})$ that asks the agent to refine it. 
For example, \texttt{"Build a cozy study with a bookshelf and a wooden desk"} is
a from-scratch query, while
$(W_{\text{text}}, W_{\text{image}}, \texttt{"Add a bookshelf beside the desk")}$ asks the agent to
add a bookshelf to an existing 3D world.
\end{myDef}

\begin{myDef}
\textbf{3D tools.} 
To construct an interactive 3D world, the agent is provided with five tools: 
\texttt{asset\_retrieve}, \texttt{asset\_add}, \texttt{asset\_rotate}, 
\texttt{asset\_translate}, and \texttt{asset\_delete}. 
These tools enable the agent to retrieve assets from the database, add assets to the scene, 
rotate assets, translate their positions, and remove assets from the 3D world, respectively. 
For example, \texttt{asset\_rotate("00001", "Bookshelf\_01", (100, 140, 0), (0,0,0), (0,90,0), 1.0)} rotates the bookshelf (located at \texttt{(100,140,0)}) $90^\circ$ around the $y$-axis while keeping its original scale.
\end{myDef}

\begin{table}[t]
\vspace{-30pt}
\centering
\caption{Illustrative example of key concepts in \method.}
\label{tab:notation}
\resizebox{\textwidth}{!}{%
\begin{tabular}{@{}c p{5.8cm} p{8.8cm} c@{}}
\toprule
\textbf{Concept} & \textbf{Description} & \textbf{Sample Format} & \textbf{Records} \\
\midrule
3D asset & The atomic block of a 3D world &
\texttt{(id, name, category, face\_count, bbox)} & 2{,}616 \\
3D world & A set of placed assets  &
\texttt{\{(m, pos[x,y,z], rot[x,y,z], sca), ...\}} & 323 \\
\midrule
\multirow{2}{*}{
\begin{tabular}{c}
Multimodal\\
query
\end{tabular}
} & \ding{172} 3D world construction&
\texttt{(textual instruction)} & \multirow{2}{*}{6{,}828} \\
 & \ding{173} 3D world refinement &
\texttt{(W\_text, W\_image, textual instruction)} & \\
\midrule
\multirow{5}{*}{3D tools}
 & \texttt{asset\_retrieve}: retrieve 3D asset &
\texttt{asset\_retrieve(name, top\_k)} & \multirow{7}{*}{5} \\
 & \texttt{asset\_add}: put a retrieved asset  &
\texttt{asset\_add(id, name, pos[x,y,z], sca)} & \\
 & \texttt{asset\_rotate}: rotate an asset &
\texttt{asset\_rotate(id, name, pos[x,y,z], rot\_origin[x,y,z], rot\_modify[x,y,z], sca)} & \\
 & \texttt{asset\_translate}: translate an asset &
\texttt{asset\_translate(id, name, pos\_origin[x,y,z], pos\_modify[x,y,z], sca)} & \\
 & \texttt{asset\_delete}: remove an asset  &
\texttt{asset\_delete(id, pos[x,y,z], name)} & \\
\bottomrule
\end{tabular}%
}
\vspace{-10pt}
\end{table}

\subsection{Problem Formulation}
\label{sec:agentic-def}

With the above definition, we now formulate the problem studied in this paper.

\begin{pro}
\textbf{Agentic 3D world construction.}
Given a multimodal query $q$, the agent interacts with the sandbox environment for $T$ turns and finally constructs an interactive 3D world, i.e., the 3D map $W_{\text{text}}^{(T)}$ and rendered multi-view images $W_{\text{image}}^{(T)}$.
At the $i$-th turn, conditioned on the query and the interaction history, the agent
generates a thought $\tau_i$ and an action $a_i$:
\begin{equation}
\{\tau_i, a_i\} = \pi_\theta\!\left(q, \{\tau_1, a_1, o_1, \ldots, \tau_{i-1}, a_{i-1}, o_{i-1}\}\right),
\end{equation}
where $\tau_i$ is the thought, $a_i$ is the action consists of one or multiple tool calls selected from our 3D tools, and the observation $o_i$ returned by the sandbox consists of the tool
response (e.g., the asset retrieval result) together with the current 3D map $W_{\text{text}}^{(i)}$ and its rendered multi-view images $W_{\text{image}}^{(i)}$.
The interaction $y=\{\tau_1, a_1, o_1, \ldots, \tau_T, a_T, o_T\}$ runs for $T$ turns and terminates when the agent issues no further
tool call, yielding the final interactive 3D world $(W_{\text{text}}^{(T)}, W_{\text{image}}^{(T)})$.
\end{pro}

\section{VWE-Bench Construction}
\label{sec:bench}

We build \bench through a collaboration between MLLMs and human annotators.
This collaboration yields high-quality 3D assets, seed 3D worlds, and multimodal queries for the vibe worlding agent task, which prior indoor 3D-scene works \citep{zhao2026scenerevis} cannot offer.
The pipeline proceeds in three stages. \textbf{Stage~1: high-quality 3D asset
synthesis}, where an artist-guided generative production line builds a large,
retrievable 3D asset set.
\textbf{Stage~2: high-quality
seed 3D world annotation}, where artists use obtained assets to construct coherent seed worlds.
\textbf{Stage~3: reverse query
synthesis}, on top of the annotated worlds, we use an MLLM to critique a seed world and leverage its feedback (i.e., indicate where and how the 3D world can be improved) to synthesize editing instructions (\emph{i.e., 3D world refinement}), or prompt an MLLM to read a seed world and write a query that reproduces it from an empty map (\emph{i.e., 3D world construction}).

\subsection{High-quality 3D Asset Synthesis}
\label{sec:asset}

To obtain a large, physically consistent asset library, we adopt an art-annotators guided asset synthesis method: the artist-annotators define the asset inventory (e.g., name and category) and control quality, while generative models (e.g., MLLMs and image-to-3D models) synthesize the reference asset images and 3D meshes. 
The line runs in three steps:
\begin{itemize}[leftmargin=*, nosep]
\item \textbf{Asset textual and image annotation.} 
Art-annotators first determine a
minimal asset inventory that covers natural open-world scenes, specifying
3,148 native concept assets with their names and categories.
For each concept asset, they further obtain an asset image through text-to-image generation using the Gemini~3.1-flash-image model.

\item \textbf{Image-to-3D generation.}
Then, each asset image is transformed directly to a 3D mesh (i.e., a \texttt{.glb} file) with an image-to-3D model.
In this paper, we use Hunyuan3D~3.1~\citep{hunyuan3d2025} to facilitate this generation process.

\item \textbf{Quality filtering and size annotation.} 
Image-to-3D generation is not always faithful to the input image. Therefore, we first filter out generated 3D assets whose geometry mesh or appearance significantly differs from the concept image (e.g., a street lamp is provided as input image, but the generated 3D mesh resembles a desk lamp).
After filtering, every surviving asset is then annotated at a unified real-world
scale, marking its center and half-extent along each of the $x/y/z$ axes to define
a native bounding box.
\end{itemize}

\begin{figure}[t]
\vspace{-30pt}
\centering
\includegraphics[width=\linewidth]{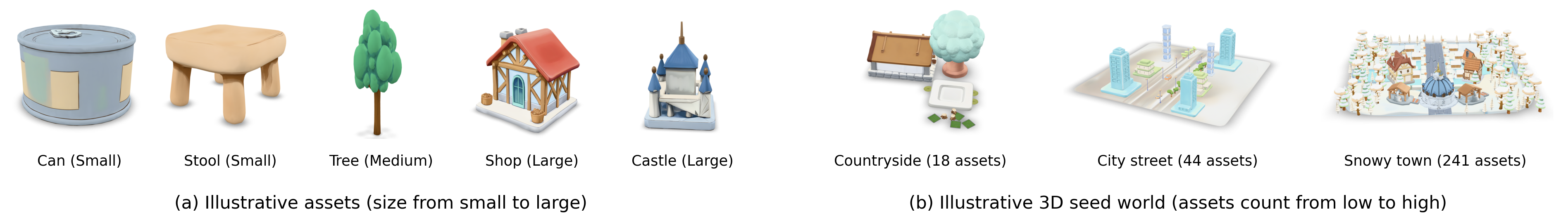}
\caption{Illustrative \bench data. \textbf{(a)} Asset examples ordered by
physical size (small to large). \textbf{(b)} Seed 3D world examples ordered
by complexity (low to high).}
\label{fig:assets}
\end{figure}

\begin{figure}[t]
\vspace{-10pt}
\centering
\includegraphics[width=\linewidth]{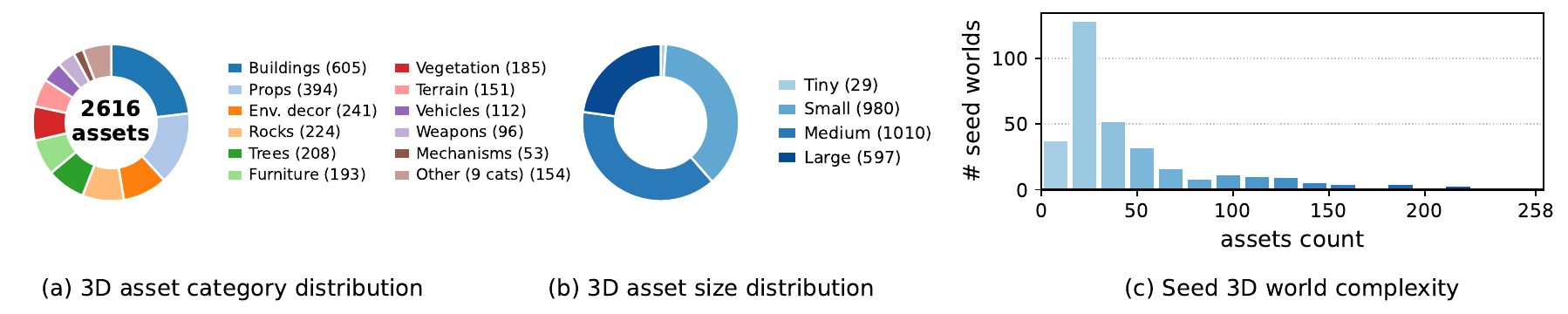}
\caption{\bench distributions. \textbf{(a)} assets across the 20 semantic categories;
\textbf{(b)} assets are annotated physical-size class; 
\textbf{(c)} the seed world complexity varies.}
\label{fig:assetstats}
\vspace{-10pt}
\end{figure}

This synthesis pipeline yields 2,616 high-quality assets.
We show illustrative asset example in Figure~\ref{fig:assets}(a) and the asset category distribution in Figure~\ref{fig:assetstats}(a).
As can be seen, the constructed 3D asset span 20 semantic categories, from small-size props to large-size buildings and terrain.
We also conduct a preliminary size annotation for all assets to enrich the asset annotation schema, whose distribution is reported in Figure~\ref{fig:assetstats}(b).
In addition, each asset also carries a semantic description, color, native size and different face-count.

\subsection{Seed 3D World Annotation}
\label{sec:seed}

To validate the practical usability of the collected assets, we ask professional art-annotators to construct 323 seed 3D worlds directly from them, jointly covering the entire synthesized 3D asset library.
Specifically, annotators are instructed to create coarse but functional 3D worlds rather than highly polished environments, as these seed worlds serve as starting points for synthesizing subsequent world refinement queries that further refine and enrich the scenes.
We provide illustrative examples of the annotated seed 3D worlds in Figure~\ref{fig:assets}(b), ordered by the number of placed assets from low to high. 
In addition, these seed worlds also exhibit diverse levels of complexity, ranging from 8 to 258 placed assets, as shown in Figure~\ref{fig:assetstats}(c).

\subsection{Reverse Multimodal Query Synthesis}
\label{sec:query}

Using the constructed 3D assets and the seed worlds, we reverse-synthesize two types of
queries (i.e., \emph{3D world construction} and \emph{3D world refinement}). 
We now introduce the query synthesis pipeline.

\textbf{3D world construction.} 
This agentic task requires the agent to construct a 3D world from scratch on an empty map based solely on a textual description.
To synthesize such queries, we prompt an MLLM to
read a seed world and reverse-generate a request describing how to build it.
Depending on how much detail the request exposes, we obtain three tiers of
increasing specificity. 
\begin{itemize}[leftmargin=*, nosep]
\item \textbf{Theme only.} 
This type of query consists of only a high-level theme (e.g., ``an eerie wasteland shrine''), which specifies the desired atmosphere while leaving the contents and layout largely unspecified.
This setting evaluates whether the agent can autonomously retrieve appropriate assets and synthesize a coherent world from minimal guidance.

\item \textbf{Theme + elements.} 
This type of query provides a theme together with a coarse list of required elements and quantities, constraining what should appear without prescribing their arrangement. 
This setting evaluates whether the agent can generate a plausible spatial layout considering multiple constraints and requirements.

\item \textbf{Full blueprint.} 
We also construct the query that describes both the required elements and their intended spatial organization, serving as a near-complete world specification. 
This setting evaluates whether the agent can faithfully translate detailed instructions into an accurate 3D world.

\item \textbf{Distractor.} 
We additionally include a small set of distractor queries that embed an
infeasible sub-request, e.g., an asset absent from the library, a physics violation, an
internal contradiction, or a hopelessly under-specified ask, to validate whether the
agent clarifies and proposes alternatives rather than blindly complying.
\end{itemize}

As these are open-ended queries with no single ground-truth world, they are all \emph{unverified} and will be evaluated by an MLLM judge based on our carefully designed rubric.

\textbf{3D world refinement.}
In this task, the agent is given a textual instruction together with an existing 3D world (e.g., its 3D map and rendered 3D images) and should perform the necessary modifications to achieve the desired changes.
We synthesize such queries in two ways:
\begin{itemize}[leftmargin=*, nosep]
\item \textbf{Asset perturbation.} We directly perturb the assets of a seed world (e.g., deleting or moving assets) and let an MLLM observe the change and describe it as an instruction:
  \begin{itemize}[leftmargin=1.5em, nosep]
  \item \textbf{Asset-level edit (precise).} 
These queries describe fine-grained asset-level editing intents, such as adding, deleting, translating, or rotating specific assets.
To synthesize such queries, we first apply controlled perturbations to a seed world and then prompt an MLLM to precisely and unambiguously describe the induced changes.
Since the perturbation process explicitly defines the desired modification, each synthesized query can be verified via a ground-truth map.
% For instance, if assets $a$ and $b$ exist in the original world but are removed during perturbation, the resulting instruction ``remove assets $a$ and $b$'' can be directly verified by checking their non-existence in the final 3D world map.
  
\item \textbf{Asset-level edit (fuzzy).}
These queries represent the same types of asset-level editing intents but are expressed with vague and underspecified instructions, reflecting the ambiguity commonly observed in real-world user requests.
Instead of specifying exact targets or modification amounts, users may provide high-level descriptions such as ``tidy up the trees a bit''.

  \end{itemize}
\item \textbf{MLLM-as-critic.} 
We also leverage an MLLM to critique the seed world from a holistic scene perspective, rather than focusing solely on individual assets, and synthesize queries describing how the world could be improved:
  \begin{itemize}[leftmargin=1.5em, nosep]
  \item \textbf{Scene critique.} 
  The query voices a shortcoming of the current world (e.g., it feels empty, cluttered, or unbalanced) and asks the agent to fix it.
  
  \item \textbf{Scene guidance.}
The query provides high-level directional guidance toward a desired effect (e.g., ``make it feel more lively'') by suggesting how the scene could be improved, rather than merely criticizing its current state.
  
  \item \textbf{Scene restatement.} 
A restatement of the desired end state, mimicking real-world scenarios where users reiterate their expectations without specifying concrete editing operations, leaving the agent to realize the goal through appropriate edits.
  
\item \textbf{Complex description.}
A complex query that describes the scene transformation, involving multiple coordinated edits or rich requirements.
Such queries are designed to stress-test the agent's ability to understand, decompose, and execute highly complex user requests.

  \end{itemize}
\end{itemize}

\begin{wraptable}{r}{0.55\textwidth}
\vspace{-5pt}
\vspace{-\baselineskip}
\centering
\caption{The query statistics in \bench.}
\label{tab:stats}
\resizebox{0.53\textwidth}{!}{%
\begin{tabular}{llccc}
\toprule
Query type & Sub-type & Verified & Unverified & Count \\
\midrule
\multirow{4}{*}{3D world construction}
 & Theme only              &            & \ding{51} & 322 \\
 & Theme + elements        &            & \ding{51} & 620 \\
 & Full blueprint          &            & \ding{51} & 302 \\
 & Distractor              &            & \ding{51} & 120 \\
\midrule
\multirow{6}{*}{3D world refinement}
 & Asset-level edit (precise)   & \ding{51}  &           & 1{,}710 \\
 & Asset-level edit (fuzzy)     &            & \ding{51} & 1{,}462 \\
 & Scene critique          &            & \ding{51} & 553 \\
 & Scene guidance          &            & \ding{51} & 757 \\
 & Scene restatement       &            & \ding{51} & 477 \\
 & Complex description     &            & \ding{51} & 505 \\
\midrule
\multicolumn{2}{l}{Total}  & 1{,}710    & 5{,}118   & 6{,}828 \\
\bottomrule
\end{tabular}%
}
\vspace{-\baselineskip}
\end{wraptable}
Overall, this process yields six types of user queries. 
Only the asset-level edit (precise) are \emph{Verified}, their instructions are precise enough that the resulting world can be compared against the ground-truth map. 
The remaining five are open-ended and designed to simulate real-world users' fuzzy expression. They are treated as \emph{Unverified} query and will be evaluated with our carefully designed rubrics.

\textbf{Statistics of \bench.}
Finally, our art-annotators help review and filter out low-quality synthesized queries, resulting in the final benchmark.
We summarize the benchmark statistics in Table~\ref{tab:stats}.
In total, \bench comprises 2,616 high-quality assets, 323 seed 3D worlds, and 6,828 reverse-synthesized queries, including 1,364 3D world construction queries and 5,464 3D world refinement queries.
% By supervision signal, 1,710 are
% \emph{Verified} precise-edit queries, where each paired with a ground-truth world
% $W^\ast$ and rule-checkable criteria, while the remaining 5,118 are \emph{Unverified} and scored by an LLM judge against rubric criteria.

\section{VibeWorlding-Gym}
\label{sec:framework}

In this paper, our goal is to investigate how vibe worlding agents can improve through agent post-training.
To this end, we build \gym to support scalable agentic RL training.
It consists of i) a stable 3D sandbox that the agent can interact with, ii) a reliable verifier that evaluates agentic 3D world construction task end-to-end, and iii) a unified post-training framework.

\subsection{3D Sandbox Construction}
\label{sec:sandbox}

\textbf{Unified 3D Tools.}
As shown in Table~\ref{tab:notation}, we build five unified tools which can be used to construct and edit a 3D world: \emph{asset retrieve}, \emph{asset add}, \emph{asset delete}, \emph{asset rotate}, and \emph{asset translate}.
Among them, \emph{asset retrieve} maps a natural-language intent to placeable asset candidates, while the remaining tools manipulate assets in the 3D world according to the provided parameters, including adding, deleting, rotating, or translating assets.
In this paper, we build the retriever upon Qwen3-Embedding-4B by first constructing a fully synthetic positive--negative pair dataset and then training it with the InfoNCE loss for asset retrieval and ranking.
Because the retriever shares the asset library's \texttt{type\_id} system, every retrieved candidate is guaranteed to be placeable.
We defer the full retrieval pipeline and training objective to Appendix~\ref{app:retrieval}.

\textbf{Unified 3D Rendering.}
We use Blender as a unified rendering service.
Specifically, after the agent edits the 3D world in a turn, the service renders the agent-modified 3D world and returns the result to the agent.
At every turn, we provide the agent with rendered images of the 3D world captured from five fixed camera viewpoints.

\subsection{Dual-Constraint Verifier}
\label{sec:verifier}

A valid 3D world should satisfy two constraints simultaneously: 1) it should be \emph{physically feasible}, and 2) it should fulfill the \emph{user's intention}.
Built on this, we propose a dual-constraint verifier that integrates physical feasibility verification with a rubric-based LLM judge to evaluate both physical validity and alignment with user intent.

\paragraph{Physical Feasibility Verification.} A valid 3D world should first respect basic physical-reality constraints. 
% For instance, two objects cannot occupy the same space, and an object cannot float unsupported in mid-air.
We verify these constraints through Python-based geometric checks:
\begin{itemize}[leftmargin=*, nosep]
\item \textbf{Collision.} The assets in 3D world should not collide, interpenetrate, or clip through one another.
\item \textbf{Height.} 
The assets in 3D world should satisfy physical support constraints by being grounded on the terrain or valid surfaces, with no unsupported floating.
\end{itemize}

\paragraph{Intent Fulfillment Verification.} 
In real-world interaction, the user's intent is open-ended and varied, so we further verify whether the agent's response and the constructed 3D world actually meet that intent.
We build an MLLM-based judge that evaluates whether an agent-constructed 3D world is acceptable along the following four aspects:
\begin{itemize}[leftmargin=*, nosep]
\item \textbf{Ecological plausibility.} Whether the agent assembles a globally coherent world (e.g., stylistically consistent and free of out-of-context objects), rather than a physically valid but incongruous scene.
\item \textbf{3D understanding.} Whether the agent correctly grasps what the user asks for, e.g., the entities and layout described in the query, and reflects them in the world.
\item \textbf{3D reasoning.} Whether the agent has understood the user's spatial intent, and could also correctly use appropriate 3D tools to modify the 3D world, rather than just understanding the goal yet editing the 3D world incorrectly.
\item \textbf{Retrieval plausibility.} Whether the agent can perform the initial asset retrieval step by selecting semantically appropriate assets, since the retrieval errors can propagate and affect all downstream 3D world editing.
\end{itemize}

\textbf{Multi-Dimension Reward Calculating.}
For \textbf{Unverified} queries, we apply the two aforementioned parts in sequence, i.e., Physical Feasibility Verification first, and only a world that passes it proceeds to the Rubric-based Judge.
Then we count the world as correct only when all of the above dimensions pass.
For \textbf{Verified} queries, a ground-truth map exists, so we directly compare the agent-edited world against it and score by the proportion of assets that are correctly modified.

\subsection{Joint Multimodal RL Post-Training}
\label{sec:posttrain}

\paragraph{Cold-start Data Synthesis for SFT.} We synthesize SFT trajectories by the query type.
For \textbf{Unverified} queries, which admit no single ground-truth world, we prompt Gemini~3.1-pro to construct the 3D world over multiple attempts, keep only the high-quality outcomes as judged by our verifier.
Then, we reverse-prompt it to rewrite a complete, coherent reasoning trajectory that leads to the retained 3D world.
For \textbf{Verified} queries, the ground-truth map is available, so we directly reverse-synthesize the agent's reasoning trajectory from it.
We then conduct full-parameter supervised fine-tuning on these cold-start trajectories, endowing the MLLM with the basic abilities of 3D tool use and multi-turn reasoning.

\paragraph{Joint Multimodal RL.} Unlike prior work that trains on a single modality, we let the agent learn jointly from both pure-text queries (constructing a world from scratch) and multimodal queries (refining an existing world given its renders).
Starting from the cold-started agent policy and using our constructed dual-constraint verifier, we use GRPO~\citep{shao2024deepseekmath} for agentic RL training.
We optimize with the outcome-based reward defined in Section~\ref{sec:verifier}: the reward is binary ($0/1$) for Unverified queries, whereas for Verified queries it is a score in $[0,1]$ given by the proportion of assets the agent modifies correctly.
Relying on this outcome-based reward design, we avoid reward hacking from hand-crafted intermediate shaping.

\section{Experimental Setup}
\label{sec:setup}

\subsection{Dataset Statistics}
\label{sec:dataset-stats}

\begin{wraptable}{r}{0.5\textwidth}
\vspace{-\baselineskip}
\centering
\caption{The training and test statistics of VWE-Bench.}
\label{tab:split}
\resizebox{0.5\textwidth}{!}{%
\begin{tabular}{lccc}
\toprule
 & \multicolumn{2}{c}{Training} & \\
\cmidrule(lr){2-3}
Query type & SFT cold-start & RL & Test \\
\midrule
3D world construction & 1{,}129 & 189 & 46 \\
3D world refinement   & 4{,}438 & 818 & 208 \\
\midrule
Total & 5{,}567 & 1{,}007 & 254 \\
\bottomrule
\end{tabular}%
}
\vspace{-\baselineskip}
\end{wraptable}

To guarantee that \bench measures genuine generalization rather than memorization, we split the dataset into training (for SFT Cold-start and RL) and testing sets with completely disjoint seed 3D worlds, as shown in Table~\ref{tab:split}.
In dataset partition, we also preserve query type proportions in Table~\ref{tab:stats} to make sure the agent can learn different policies.
During training, we conduct cold-start data synthesis using SFT query set, whereas the RL query set is directly used for joint multimodal RL.

\subsection{Metric and Evaluation Protocol}
\label{sec:metric}

\textbf{Auto Evaluation.} \bench provides automatic evaluation using an MLLM (we use Gemini 3.5-flash as our verifier backbone in this paper): every constructed 3D world can be scored by the dual-constraint verifier of Section~\ref{sec:verifier}.
And we report \textbf{Pass@1}, the fraction of queries solved by a single rollout.
An \emph{Unverified} query counts as passed only when all verifier dimensions pass.
For a \emph{Verified} query we use the rule-based score (the proportion of correctly modified assets), so Verified Pass@1 is the mean of this score.

\textbf{Human Evaluation.} 
To complement the automatic metric, we run a blind human annotation. 
For each model, we present that model's outputs to art annotators.
Each sample is presented anonymously, where annotators see only the user query, five multi-view screenshots of the initial 3D world, five of the final 3D world, and the agent's final natural-language response, without access to the underlying model identity.
The evaluation spans both the constructed 3D world and the agent's final response.
\textbf{For the constructed 3D world:} (i) \emph{user intent fulfillment}, (ii) \emph{ecological plausibility} on a $0/1$ judgement, and (iii) \emph{physical feasibility} (the automatic collision result is shown for reference). 
\textbf{For the final response:} (iv) \emph{factual hallucination}, i.e.\ whether the response claims a 3D-world edit that was not actually performed, and (v) \emph{response intelligence} on a $1$--$5$ scale, i.e.\ whether the response appropriately clarifies under-specified or unreasonable requests.

\textbf{Evaluation Quality Analysis.}
We validate the automatic verifier against the blind human labels on the same samples, dimension by dimension. The verifier is well-aligned with human judgement: at the aspect level it reaches Cohen's $\kappa=0.54$ on ecological plausibility ($93.6\%$ agreement) and $\kappa=0.53$ on intent fulfillment ($78.0\%$), and at the case level it agrees with the holistic human Pass@1 $83.4\%$ of the time ($\kappa=0.54$). Most importantly, at the system level the two rankings are highly consistent: verifier reward and human intent-fulfillment rate correlate at Spearman's $\rho=0.88$ across all evaluated systems ($8$ frontier MLLMs, $3$ agent-scaffold frameworks, and $6$ open backbones and their post-trained variants), so the verifier preserves the model ordering that we ultimately care about.
We provide more details in Appendix~\ref{app:human-eval}.

\subsection{Baselines}
\label{sec:baselines}

All methods operate over the same sandbox tools, asset library, and verifier for a fair comparison, and fall into three families.
\textbf{(i) Frontier MLLMs} used zero-shot as the agent policy, spanning closed and open models from $8$B to $2.8$T parameters: GPT-5.5, Gemini~3.5-flash, Gemini~3.1-pro, Claude-Opus-4.8, Kimi-K3 and Qwen3.8-Max; we further include the open Qwen3-VL-8B / 30B-A3B base models (Table~\ref{tab:bytype}).
\textbf{(ii) Agent-scaffold baselines}: we adapt three representative \emph{training-free} 3D-scene agent frameworks---SceneWeaver~\citep{yang2025sceneweaver}, SAGE~\citep{xia2026sage}, and SceneAssistant~\citep{luo2026sceneassistant}.
Specifically, we re-implement their planning and self-refinement workflows on top of our unified sandbox tools with GPT-5.5 as a fixed backbone, isolating the effect of agent scaffolding from that of our post-training.
\textbf{(iii) Ours}: VibeWorlder-8B and VibeWorlder-30B-A3B, each in its cold-start SFT and RL-post-trained variants, with the corresponding Qwen3-VL base models as references.

\subsection{Implementation Details}
\label{sec:impl}

For \textbf{Cold-start}, we run full-parameter supervised fine-tuning for $2$ epochs with a maximum sequence length of $122{,}880$ tokens, and learning rate $2\times10^{-5}$ on reward-filtered cold-start trajectories.
For \textbf{RL}, we optimize with GRPO~\citep{shao2024deepseekmath} for one epoch at learning rate $5\times10^{-7}$, KL coefficient $0.05$, and no entropy bonus.
Each prompt draws a group of $8$ rollouts.
For \textbf{Sandbox \& Vision Pixel}, the Blender service returns five $1280\times720$ renders as multimodal feedback every turn.
The RL reward verifier is served by Gemini 3.5-flash.
The 8B models are trained on a single node with 8$\times$NVIDIA H20 GPUs, while the 30B-A3B models are trained on three GPU nodes.

\section{Result Analysis}
\label{sec:results}

\subsection{Main Results}
\label{sec:main-results}

\begin{figure}[t]
\vspace{-25pt}
\centering
\includegraphics[width=\linewidth]{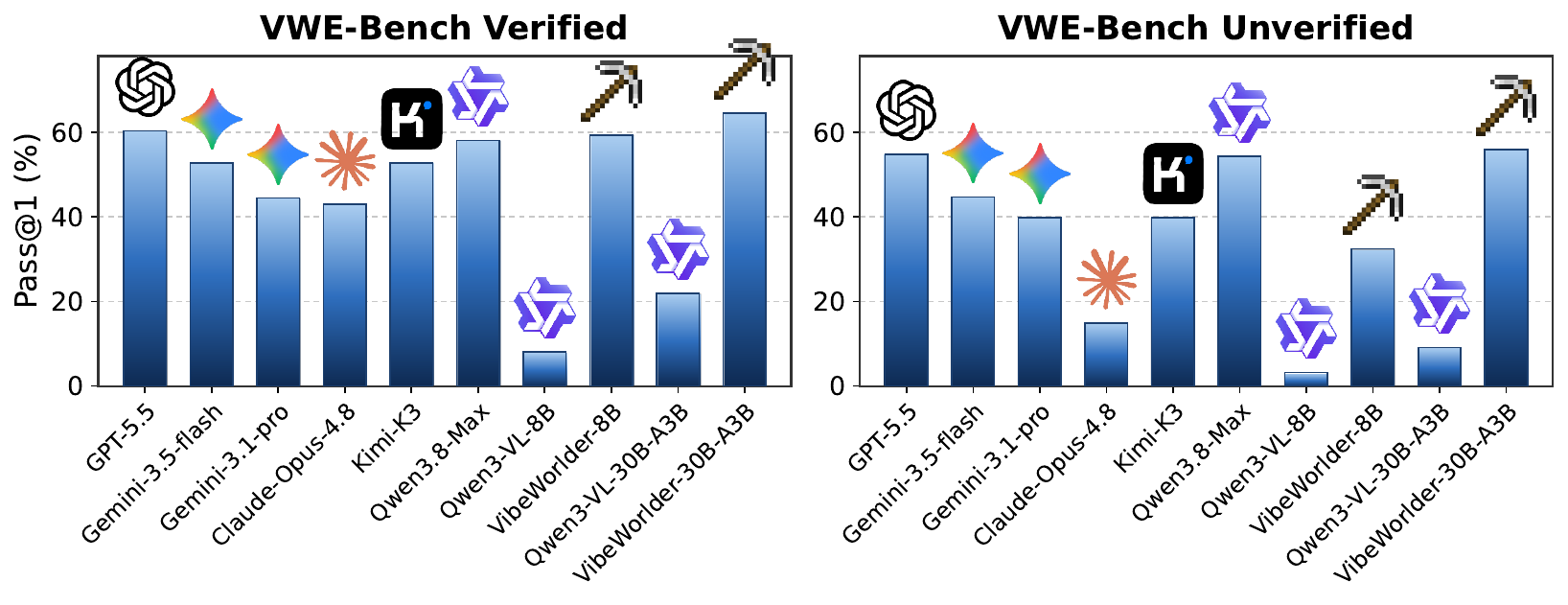}
\caption{Overall Pass@1 on \bench across models on Verified and Unverified tasks.}
\label{fig:main-bar}

\end{figure}

\textbf{Current MLLMs are far from solving the task.}
As shown in Figure~\ref{fig:main-bar}, even the strongest frontier models (e.g., GPT-5.5 and Qwen3.8-Max) achieve less than $\sim60\%$ overall Pass@1, indicating that end-to-end 3D world construction remains far from solved. In contrast, untrained open-source models perform substantially worse, with Qwen3-VL-8B and Qwen3-VL-30B-A3B achieving only $5.3\%$ and $13.6\%$ Pass@1, respectively.

\textbf{Multimodal RL post-training closes the gap to the frontier.}
As can be seen, cold-start SFT and joint multimodal RL together lift the open backbones substantially: the 30B-A3B model climbs from $13.6\%$ (base) to $34.5\%$ (SFT) and then to $59.3\%$ after RL, while the 8B model reaches $41.4\%$.
Notably, VibeWorlder-8B ($41.4\%$ overall) already matches Gemini~3.1-pro ($42.7\%$) and \emph{surpasses} it on the Verified track ($59.3$ vs.\ $44.4\%$), where precise, rule-checkable editing is required.
Scaling up, our flagship VibeWorlder-30B-A3B attains the \emph{best overall Pass@1 across all evaluated models} ($59.3\%$), outperforming the strongest closed-source frontiers GPT-5.5 ($57.3\%$) and Qwen3.8-Max ($56.9\%$); its lead is most pronounced on the Verified track ($64.5\%$ vs.\ GPT-5.5's $60.4\%$), confirming that agentic RL with reliable rewards is especially effective at instilling the precise, collision-aware editing that frontier models still lack.

\begin{table}[t]
\vspace{-20pt}
\centering
\caption{Pass@1 (\%) by query sub-type on the Test split. 
For 3D world construction queries, the sub-types are Theme-only, Theme+elements (Th+El), Full-blueprint (Blupr), and Distractor (Distr). 
For 3D world refinement queries, the sub-types include asset edit (Precise/Fuzzy), Scene critique (Crit.), Scene guidance (Guid.), Scene restatement (Rest.), and Complex description (Cplx). 
The abbreviations in parentheses correspond to the column names in the table. 
``Overall'' denotes the mean reward over all query types.
``Human'' denotes the holistic pass rate from a blind human study per model (a sample passes only if human annotators judge it physically feasible, intent-fulfilling, ecologically plausible, and free of response hallucination) shown in Appendix~\ref{app:human-eval}.
Agent-scaffold baselines (SceneWeaver, SAGE, SceneAssistant) use GPT-5.5 as the backbone.}

\label{tab:bytype}
\resizebox{\textwidth}{!}{%
\begin{tabular}{l cccc cccccc cc}
\toprule
 & \multicolumn{4}{c}{3D world construction} & \multicolumn{6}{c}{3D world refinement} & & \\
\cmidrule(lr){2-5}\cmidrule(lr){6-11}
Model & Theme & Th+El & Blupr & Distr & Precise & Fuzzy & Crit. & Guid. & Rest. & Cplx & Overall & Human \\
\midrule
Gemini~3.1-pro       & 8.3 & 18.2 & 20.0 & 50.0 & 44.4 & 47.5 & 57.1 & 48.0 & 50.0 & 33.3 & 42.7 & 13.3 \\
Gemini~3.5-flash     & 66.7 & 31.8 & 30.0 & 50.0 & 52.8 & 52.5 & 52.4 & 48.0 & 28.6 & 28.6 & 47.4 & 43.3 \\
GPT-5.5              & 75.0 & 50.0 & 70.0 & 50.0 & 60.4 & 55.7 & 57.1 & 56.0 & 50.0 & 38.1 & 57.3 & 33.3 \\
Claude-Opus-4.8      & 25.0 & 4.5 & 20.0 & 0.0 & 42.9 & 23.0 & 4.8 & 20.0 & 14.3 & 0.0 & 24.3 & 13.3 \\
Kimi-K3              & 41.7 & 18.2 & 20.0 & 100.0 & 52.8 & 54.1 & 52.4 & 28.0 & 35.7 & 28.6 & 44.5 & 26.7 \\
Qwen3.8-Max          & 50.0 & 31.8 & 40.0 & 100.0 & 58.1 & 62.3 & 57.1 & 60.0 & 64.3 & 42.9 & 56.9 & 46.7 \\
\midrule
SceneWeaver          & 0.0 & 9.1 & 20.0 & 0.0 & 1.3 & 21.3 & 23.8 & 36.0 & 28.6 & 14.3 & 16.3 & 23.1 \\
SAGE                 & 8.3 & 22.7 & 20.0 & 0.0 & 2.3 & 16.4 & 23.8 & 20.0 & 0.0 & 4.8 & 13.6 & 22.2 \\
SceneAssistant       & 8.3 & 22.7 & 20.0 & 0.0 & 12.9 & 14.8 & 14.3 & 12.0 & 7.1 & 0.0 & 13.8 & 11.1 \\
\midrule
Qwen3-VL-8B                & 0.0 & 0.0  & 0.0  & 0.0 & 8.1  & 4.9  & 9.5  & 4.0  & 0.0  & 0.0  & 5.3 & 3.4 \\
VibeWorlder-8B-SFT       & 0.0 & 0.0  & 0.0  & 0.0 & 17.9 & 14.8 & 23.8 & 12.0 & 7.1  & 0.0  & 12.0 & 6.7 \\
VibeWorlder-8B        & 0.0 & 18.2 & 0.0  & 0.0 & 59.3 & 42.6 & 57.1 & 44.0 & 28.6 & 19.0 & 41.4 & 30.0 \\
Qwen3-VL-30B-A3B           & 0.0 & 0.0  & 0.0  & 0.0 & 22.0 & 13.3 & 20.0 & 12.0 & 7.1  & 4.8  & 13.6 & 13.3 \\
VibeWorlder-30B-A3B-SFT  & 8.3 & 0.0  & 10.0 & 0.0 & 40.4 & 37.7 & 52.4 & 48.0 & 21.4 & 25.0 & 34.5 & 20.0 \\
VibeWorlder-30B-A3B   & 58.3 & 31.8 & 60.0 & 50.0 & 64.5 & 62.3 & 57.1 & 64.0 & 64.3 & 42.9 & 59.3 & 47.2 \\
\bottomrule
\end{tabular}%
}
\end{table}

\textbf{Where Do MLLMs struggle?}
Breaking down Pass@1 across the ten query sub-types (Table~\ref{tab:bytype}) reveals that \emph{3D world construction} from scratch remains challenging for all MLLMs, as it requires jointly solving asset retrieval, 3D spatial reasoning, and physically plausible object placement while avoiding collisions.
In contrast, \emph{3D world refinement} queries better differentiate model capabilities. 
Among them, \emph{Complex Description} and \emph{Scene Restatement} are the most difficult, as they require interpreting long, underspecified real-world instructions; even frontier models achieve below $50\%$ Pass@1. Nevertheless, RL consistently improves performance across all refinement sub-types.

\begin{figure}[t]
\centering
\includegraphics[width=\linewidth]{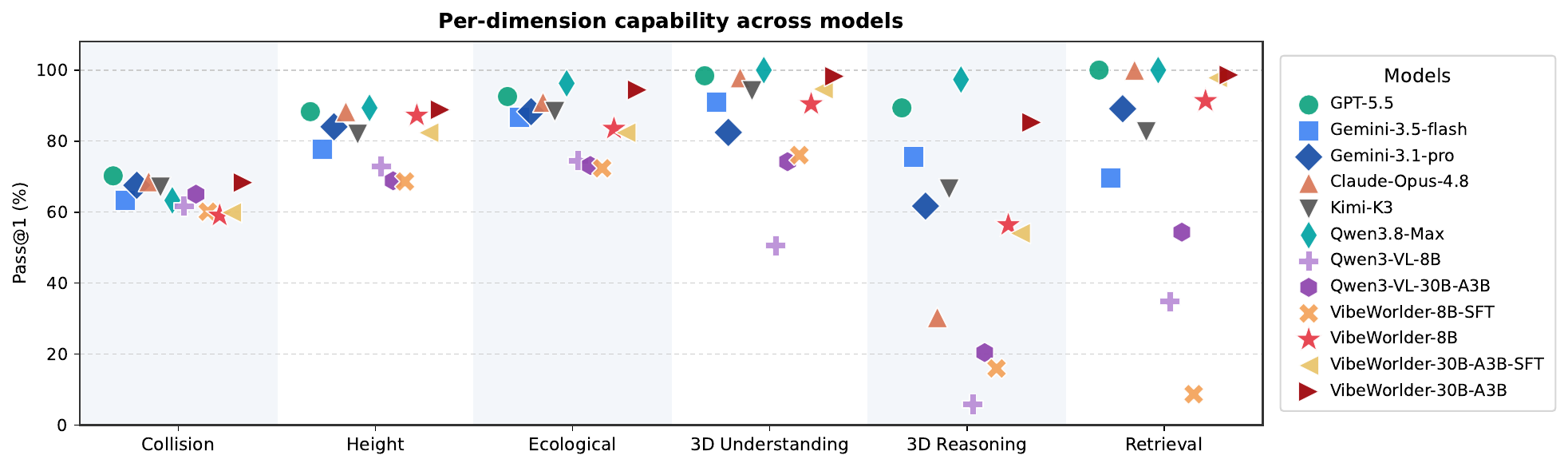}
\caption{Per-dimension pass rate across existing MLLMs, revealing which capabilities bottleneck.
Each of the six capabilities (collision-freeness, height plausibility, ecological plausibility, 3D understanding, 3D reasoning, and retrieval plausibility) is quantified by our dual-constraint verifier.}
\label{fig:bydim}
\end{figure}

\textbf{Which Capability is the Bottleneck?}
We further analyze performance across the six capability dimensions to identify the primary bottleneck, as shown in Figure~\ref{fig:bydim}. \emph{Collision} remains the weakest capability for all models, with scores of only $59$--$68\%$, even after RL training.
This makes collision-free 3D spatial editing the key unresolved challenge.
In contrast, RL substantially improves \emph{3D reasoning}, increasing performance from $6$--$20\%$ (base backbones) to $56$--$85\%$; our VibeWorlder-30B-A3B reaches $85\%$, surpassing Gemini~3.1-pro ($62\%$) and approaching the strongest frontiers.
RL also boosts \emph{retrieval} usage to $91$--$99\%$, surpassing the frontier model, while consistently improving \emph{3D understanding} ($90$--$98\%$) and \emph{ecological plausibility} ($84$--$94\%$).
Overall, RL enables models to better understand the scene and retrieve appropriate assets, but accurately placing them without introducing collisions remains the primary bottleneck.

\begin{figure}[t]
\vspace{-25pt}
\centering
\includegraphics[width=\linewidth]{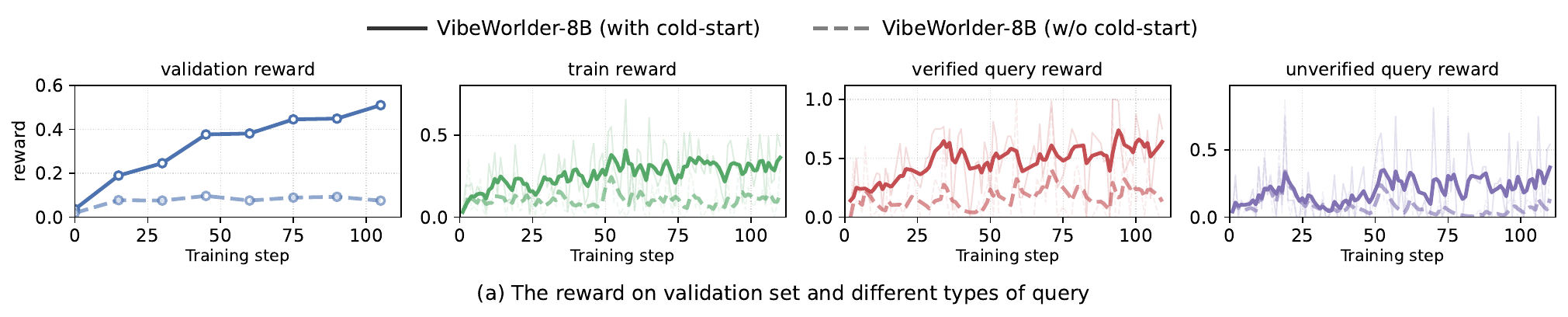}\\[1pt]
\includegraphics[width=\linewidth]{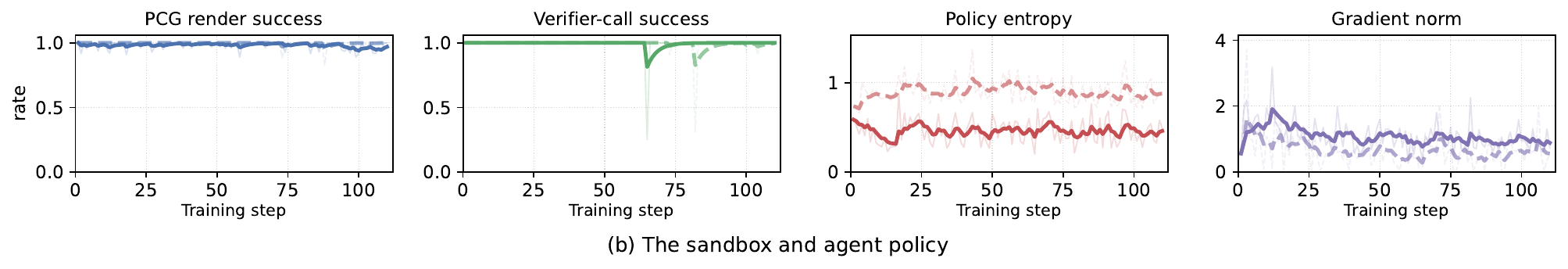}\\[1pt]
\includegraphics[width=\linewidth]{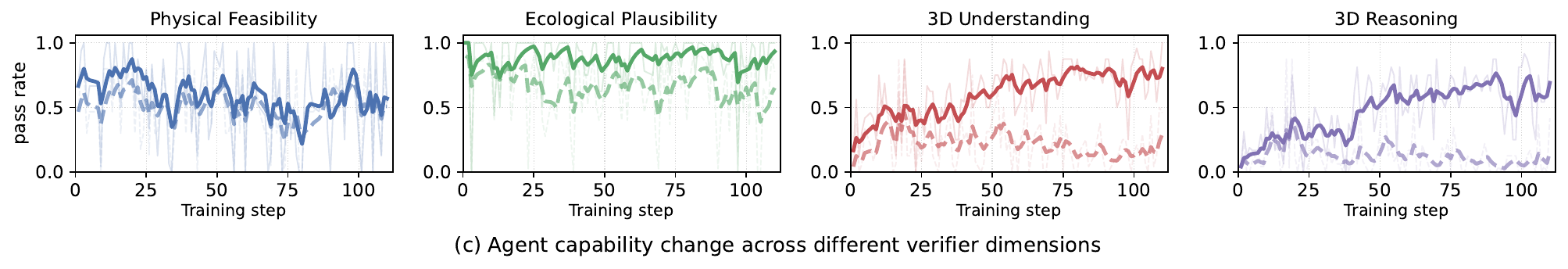}
\caption{The Metric dynamics of the VibeWorlder-8B in multimodal RL training process. 
Solid curves are the flagship run with cold-start (i.e., RL from the cold-start policy), and dashed curves are the \emph{w/o} cold-start ablation (i.e., RL from the base backbone).}
\label{fig:rlcurve}
\end{figure}

\subsection{Agent Capability Improvement in Multimodal RL Post-Training}
\label{sec:rl-curves}

To understand how the agent's capabilities emerge and what each training stage contributes, we track VibeWorlder-8B over the multimodal RL post-training process and additionally run an ablation that applies RL directly to the base backbone (\emph{w/o} cold-start; dashed curves in Figure~\ref{fig:rlcurve}). 
We organize our key insights around the following two questions.

\textbf{How do SFT and Multimodal RL Boost MLLMs, respectively?}
The two stages improve complementary capability axes (Figure~\ref{fig:rlcurve}(c)).
Cold-start SFT is chiefly responsible for the foundational physical and ecological competence: at the start of RL the agent already hovers around $\sim\!0.6$ physical feasibility and $\sim\!0.9$ ecological plausibility, both of which then stay essentially flat throughout RL, leaving little headroom to exploit.
However, SFT delivers gains almost exclusively in 3D understanding (from $0.02$ to $0.17$), with limited improvement in 3D reasoning. 
Multimodal RL is what unlocks the 3D spatial capabilities: 3D understanding improves from $0.17$ to $0.80$, while 3D reasoning, the most challenging dimension and the weakest capability at initialization, rises from $0.04$ to $0.69$. 
Nevertheless, its post-RL performance remains noticeably below that of understanding, indicating that spatial reasoning is only partially unlocked by reinforcement learning.

\textbf{How does RL Boost the Agent on Verified vs.\ Unverified queries?}
As shown in Figure~\ref{fig:rlcurve}(a), we decompose the reward by query type to reveal two distinct learning regimes.
On verified queries, where rewards are derived from deterministic ground-truth checks, the learning signal is precise and reliable. As a result, performance improves both substantially and stably: the verified reward rises smoothly from $0.14$ to $0.64$, with corresponding validation reward gains that indicate genuine generalization.
On unverified queries, the reward is instead provided by an MLLM-as-judge guided by a hand-designed rubric. Although the agent still achieves substantial improvement (from $0.04$ to $0.37$), the training reward curve is considerably more unstable. This suggests that the reliability of RL improvement is ultimately bounded by reward fidelity: verifiable rewards provide a more precise and consistent learning signal than MLLM-based judgment.

\subsection{Failed Case Analysis}
\label{sec:failcase}

\begin{figure}[t]
\vspace{-25pt}
\centering
\includegraphics[width=\linewidth]{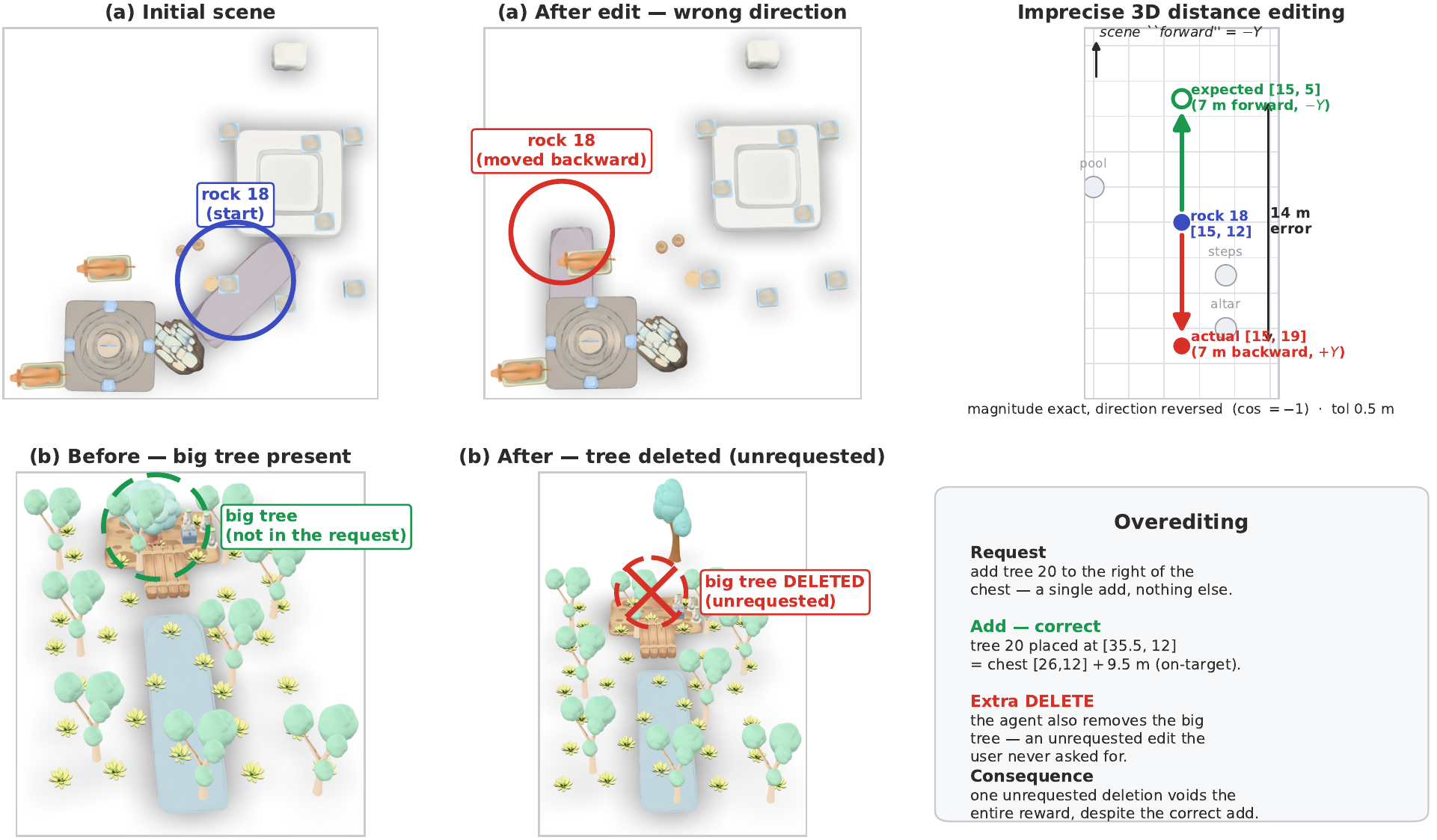}
\caption{Two dominant failure modes of frontier agents on \bench. 
\textbf{(a) Imprecise 3D distance editing:} The query requires moving rock~18 ``$7$\,m forward.'' 
The agent achieves the correct displacement magnitude but moves the object in the opposite direction, resulting in a $14$\,m positional error under a $0.5$\,m tolerance. The initial and edited renders (left and right) and coordinate visualization (center) highlight this directional inversion.
\textbf{(b) Overediting:} The query requires adding a tree beside the chest. 
The agent correctly adds the new tree but silently deletes the existing tree, introducing an unrequested modification that violates the editing constraint.
}
\label{fig:failcase}
\end{figure}

To understand what blocks the vibe worlding agent, we manually inspect the failed cases of our VibeWorlder-30B-A3B and frontier MLLMs (e.g., GPT-5.5 and Qwen3.8-Max), and identify two common and dominant failure modes that persist, shown in Figure~\ref{fig:failcase}.

\textbf{Imprecise 3D distance editing.}
The dominant failure is a gap between spatial intent and 3D spatial execution: the agent selects the right asset and the right semantic target, but emits wrong 3D coordinates.
As shown in Figure~\ref{fig:failcase}(a), the agent translates the object by exactly the requested distance ($7$,m), but along the precisely opposite direction (mapping ``forward'' to $+Y$ instead of the scene's $-Y$ axis). This failure reflects a coordinate-frame misunderstanding (cosine similarity $=-1$), rather than an error in translation magnitude.
This pattern occurs on both of the end-to-end 3D world construction and 3D world refinement tasks, indicating that precise 3D distance editing is the bottleneck.
This led to the asset collision issue illustrated in Figure~\ref{fig:failcase}(a), which is also consistent with our capability dimension analysis in Figure~\ref{fig:bydim}.

\textbf{Overediting on constructed 3D worlds.}
The second failure mode is scope violation during multi-turn refinement: the agent modifies assets the user never asked it to change.
For example, as shown in Figure~\ref{fig:failcase}(b), the agent is instructed to add a tree beside a chest. 
Although its plan correctly captures both intents and the new tree is placed at the desired location, the agent incorrectly executed a deletion operation, causing the large tree to disappear entirely.
The failure is thus not due to misunderstanding the instruction, but to weak grounding of the existing 3D world state and poor preservation of untouched elements during tool execution. 
Despite correctly identifying the requested edit, the agent over-modifies the scene by using a deletion, underscoring the value of multimodal, render-in-the-loop feedback for maintaining state awareness across interaction turns.

\subsection{Real-World CLI User Study}
\label{sec:userstudy}
\begin{figure}[t]
\vspace{-25pt}
\centering
\includegraphics[width=\linewidth]{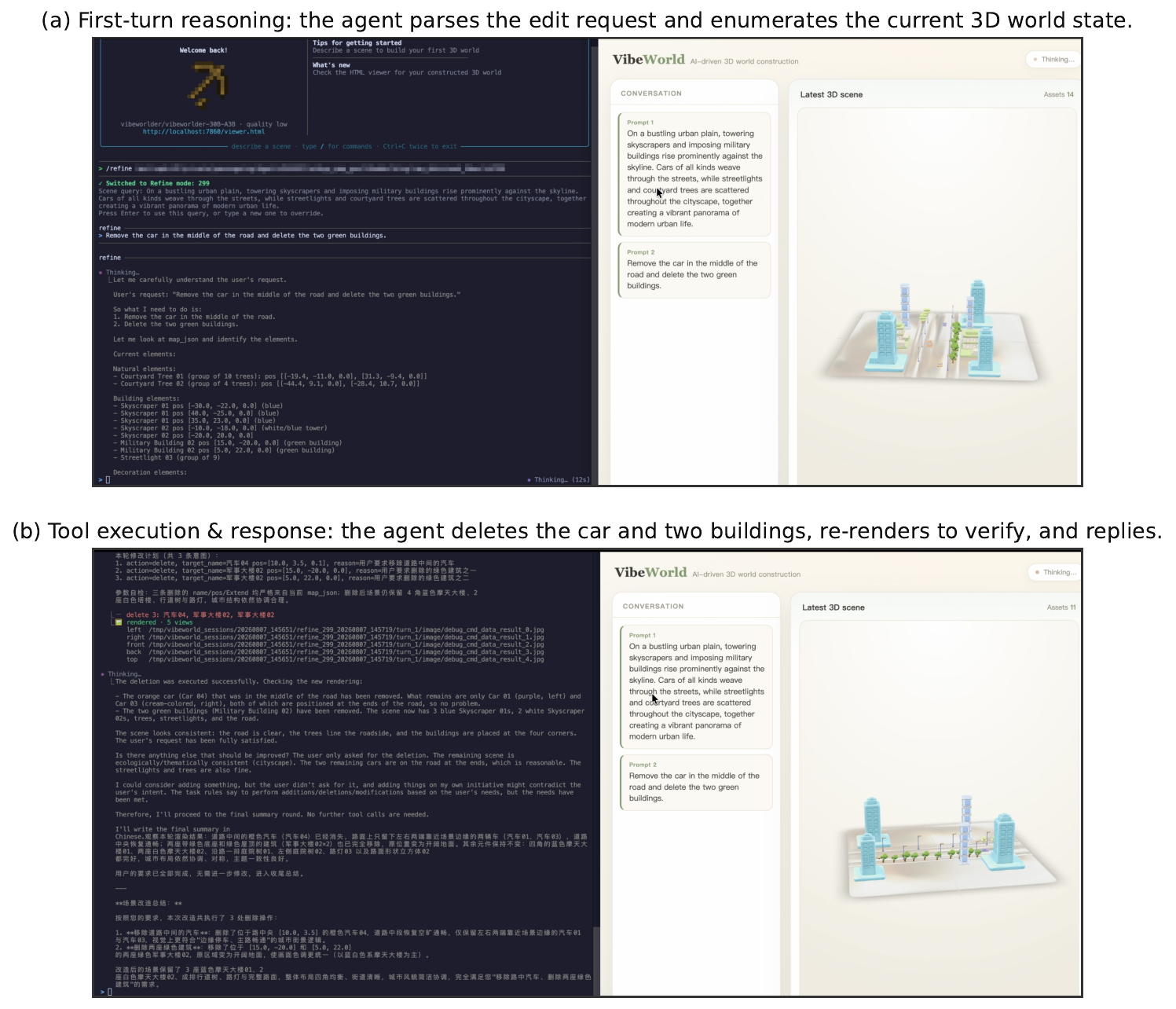}
\caption{The \textbf{VibeWorlding} CLI prototype during a multi-turn refinement session on an urban scene, with the interactive terminal (left) shown alongside the browser-based GUI viewer (right). 
(a) First-turn reasoning: after the user asks to ``remove the car in the middle of the road and delete the two green buildings,'' the agent parses the request and enumerates the current 3D world state (trees, skyscrapers, military buildings, streetlights) from the scene's 3D map. 
(b) Tool execution and response: the agent proactively invokes three \texttt{delete} tool calls, re-renders the 3D world from multiple views to verify that only the intended assets were removed, and returns a natural-language summary.
After refinement, the 3D asset count decreases from $14$ to $11$ assets accordingly.}
\label{fig:cli}
\end{figure}

To validate that our post-trained agent is usable beyond the benchmark, we wrap it in a CLI usage prototype as illustrated in Figure~\ref{fig:cli}.
Similar to vibe coding, we use an interactive terminal with a browser-based GUI viewer that renders the current 3D world after every turn.
Figure~\ref{fig:cli} shows a multi-turn refinement session on an urban scene: the user issues a natural-language edit (``remove the car in the middle of the road and delete the two green buildings''), and the agent first parses the request and grounds it against the current world state, enumerating the scene's trees, skyscrapers, military buildings, and streetlights from its \texttt{map\_json} (a).
It then emits the corresponding \texttt{delete} tool calls, re-renders the scene from multiple views, and uses the rendered images to verify that only the intended car and two green buildings were removed while every untouched element is preserved, before returning a natural-language summary to the user (b).
The GUI viewer reflects the updated world in real time.
This closed render-in-the-loop workflow mirrors the multimodal feedback used during training, and the same interface supports both 3D world construction and 3D world refinement tasks.
We release this prototype to encourage broader research efforts toward extending our system, such as developing more powerful agent harnesses, specialized skills, and other complementary components.

\section{Related Work}
\label{sec:related}

\subsection{MLLM-based Agents for 3D World Construction}
MLLM-based 3D world construction has evolved from \emph{Fixed Workflow} pipelines to \emph{Autonomous Agentic} systems, where an MLLM iteratively plans the scene, invokes external 3D tools, and revises the constructed world~\citep{ling2026scenethesis}.
For \emph{Fixed Workflow} methods, early approaches decompose world construction into a predefined sequence of stages, each handled by a specialized module or sub-agent. 
For example, SceneCraft \citep{hu2024scenecraft} synthesizes scenes as executable Blender programs under a critique-and-refine loop, while 3D-GPT \citep{sun2024gpt3d} employs multiple sub-agents to translate natural language into procedural scene-generation parameters. 
A parallel line of work integrates MLLMs with industrial game engines. 
UnrealLLM \citep{tang2025unrealllm}, LatticeWorld \citep{duan2025latticeworld}, and WorldGen \citep{wang2026worldgen} compile textual descriptions into engine-executable scene programs for generating large-scale traversable environments.
\emph{Autonomous Agentic} methods instead expose asset editing, procedural generation, and rendering as a unified tool set, allowing a single MLLM agent to autonomously decide which tool to invoke and iteratively refine the 3D world. 
For instance, SceneWeaver \citep{yang2025sceneweaver} performs planning over an extensible tool suite with self-evaluation of physical plausibility and semantic alignment. 
SAGE \citep{xia2026sage} combines generators and critics to optimize semantic, visual, and physical consistency, while SceneAssistant \citep{luo2026sceneassistant} and Vinedresser3D \citep{chi2026vinedresser3d} directly leverage rendered visual feedback to guide world construction.
Despite these advances, existing methods remain limited to idealized settings and fail to handle open-ended real-world user queries. 
Moreover, the lack of a unified open-source benchmark and training sandbox environment prevents fair comparison, scalable training, and reliable verification of 3D world construction agents, leaving progress on end-to-end agentic 3D world construction largely unexplored.

\subsection{Multimodal Reinforcement Learning}
Reinforcement learning with verifiable rewards (RLVR) has become a standard paradigm for improving long-horizon reasoning and tool use \citep{feng2025retool, ning2025deeptravel}, where rewards are provided by executable programs or structured verifiers. 
Building on this success, recent work has extended RL to multimodal agents across diverse domains. 
For example, Kimi K2.5 \citep{team2026kimi} jointly optimizes vision and language through RL, while OpenSearch-VL \citep{chen2026opensearch} trains a multimodal search agent with RL. 
In 3D world construction, recent work such as SceneReVis \citep{zhao2026scenerevis} has begun to apply multi-turn agentic RL to train world-building agents. 
However, existing approaches rely primarily on textual supervision and fail to leverage rich multimodal feedback (e.g., rendered scene images). 
To the best of our knowledge, no prior work has explored multimodal reinforcement learning for 3D world construction.

\section{Discussion}
\label{sec:conclusion}

In this paper, we present \method, a unified framework for benchmarking and training vibe worlding agents that construct interactive 3D open worlds end-to-end.
We build \bench, a benchmark of high-quality 3D assets, human-annotated seed 3D worlds, and reverse-synthesized multimodal user queries spanning both from-scratch construction and multi-turn refinement.
We further propose \gym, a unified RL infrastructure environment that unifies asset retrieval, editing, and rendering as MCP tools together with a dual-constraint verifier for scalable evaluation and multimodal RL reward.
Our comprehensive analysis reveals where current multimodal agents fall short: even frontier MLLMs remain far from solving the task, and precise, collision-free 3D editing is the dominant bottleneck.
It further clarifies what post-training can and cannot deliver: cold-start SFT establishes the foundational physical and ecological competence, multimodal RL unlocks 3D understanding and asset retrieval and only partially unlocks 3D reasoning.
The RL yields larger and more stable gains on verified queries than on unverified ones where the reward is bounded by MLLM-as-judge fidelity.
We hope these findings, together with our open-source release, help advance the frontier of end-to-end 3D world construction.

\textbf{Directions for Improvement.}
While \method takes the first step toward systematically benchmarking and training vibe worlding agents, several aspects remain open.
\begin{itemize}[leftmargin=*, nosep]
\item \textbf{Richer and more integrated 3D tools and agent skills.} 
Our current sandbox exposes a minimal set of atomic operations (retrieval, add, delete, translate, rotate). 
Future vibe worlding agents would benefit from higher-level, more composable tools and reusable 3D world construction skills, for example, directly instantiating a football field or scattering assets over a rectangular region to populate a forest, rather than assembling every scene from low-level edits.

\item \textbf{Larger-scale 3D world construction.} The most complex 3D world in \bench contains only $258$ assets, whose scale remains relatively small. Truly open-ended vibe worlding requires constructing far larger-scale 3D world environments, e.g., decomposing the 3D world construction task into multiple subtasks handled by a swarm of collaborating multimodal agents.

\item \textbf{More efficient multimodal RL methods for 3D world construction.} Our current recipe scores the entire trajectory with an outcome-based reward, which is naturally sparse for long-horizon 3D world construction. 
Conducting potential reward credit assignment (e.g., directly informing the agent of the collision at the specific turn) within our framework is a promising direction for making agentic multimodal RL more efficient and effective.

\item \textbf{More diverse data sources and user queries.} 
We build \bench using an internal crowdsourcing pipeline at Tencent, and its asset library is currently limited to a cartoon art style. 
Moreover, although our query taxonomy is derived from a user interaction study, it may not fully capture the diversity of real-world user requests and does not yet cover multimodal settings such as image-to-3D-world or video-to-3D-world task settings.
\end{itemize}

\textbf{Challenges for Vibe Worlding Agents.}
Beyond these research and engineering directions, our analysis surfaces deeper challenges that are likely to persist.
\begin{itemize}[leftmargin=*, nosep]
\item \textbf{Limited 3D spatial reasoning capability.} 
Although our experiments show that a multimodal agent can jointly leverage the 3D map and rendered world images to edit a 3D world, agents still struggle to reason precisely about spatial quantities such as distance and angular relationships, as evidenced in our capability and failed-case analysis. 
Closing this gap may ultimately require more powerful base MLLMs, achieved through approaches such as scaling up 3D world data during both pre-training and mid-training, enabling foundation models to acquire stronger spatial reasoning and world-understanding capabilities.

\item \textbf{Time-consuming 3D world rendering.} 
Vibe worlding agents rely on simulator rendering to obtain visual feedback, an inherently time-consuming step that bottlenecks efficiency and complicates data scaling for both evaluation and agentic RL training.

\item \textbf{End-to-end verification.} While our rubric-based verifier is effective within \bench, transferring it to the broader, open-ended settings of 3D world construction remains difficult. 
Vibe Worlding agent tasks remain inherently challenging to evaluate, because they are open-domain and non-verifiable tasks akin to creative writing, where assessing scene plausibility, aesthetic quality, and overall coherence remains highly subjective and difficult to automate.
\end{itemize}

\bibliography{iclr2026_conference}

@inproceedings{hu2024scenecraft,
  title={SceneCraft: An LLM Agent for Synthesizing 3D Scenes as Blender Code.},
  author={Hu, Ziniu and Iscen, Ahmet and Jain, Aashi and Kipf, Thomas and Yue, Yisong and Ross, David A and Schmid, Cordelia and Fathi, Alireza},
  booktitle={ICML},
  pages={19252--19282},
  year={2024}
}

@inproceedings{wang2026worldgen,
  title={Worldgen: From text to traversable and interactive 3d worlds},
  author={Wang, Dilin and Jung, Hyunyoung and Monnier, Tom and Sohn, Kihyuk and Zou, Chuhang and Xiang, Xiaoyu and Yeh, Yu-Ying and Liu, Di and Huang, Zixuan and Nguyen-Phuoc, Thu and others},
  booktitle={Proceedings of the IEEE/CVF Conference on Computer Vision and Pattern Recognition},
  pages={27124--27135},
  year={2026}
}

@article{duan2025latticeworld,
  title={Latticeworld: A multimodal large language model-empowered framework for interactive complex world generation},
  author={Duan, Yinglin and Zou, Zhengxia and Gu, Tongwei and Jia, Wei and Zhao, Zhan and Xu, Luyi and Liu, Xinzhu and Lin, Yenan and Jiang, Hao and Chen, Kang and others},
  journal={arXiv preprint arXiv:2509.05263},
  year={2025}
}

@inproceedings{tang2025unrealllm,
  title={Unrealllm: Towards highly controllable and interactable 3d scene generation by llm-powered procedural content generation},
  author={Tang, Song and Zhao, Kaiyong and Wang, Lei and Li, Yuliang and Liu, Xuebo and Zou, Junyi and Wang, Qiang and Chu, Xiaowen},
  booktitle={Findings of the Association for Computational Linguistics: ACL 2025},
  pages={19417--19435},
  year={2025}
}

@inproceedings{sun2024gpt3d,
  title={3d-gpt: Procedural 3d modeling with large language models},
  author={Sun, Chunyi and Han, Junlin and Deng, Weijian and Wang, Xinlong and Qin, Zishan and Gould, Stephen},
  booktitle={2025 International Conference on 3D Vision (3DV)},
  pages={1253--1263},
  year={2025},
  organization={IEEE}
}

@article{xia2026sage,
  title={Sage: Scalable agentic 3d scene generation for embodied ai},
  author={Xia, Hongchi and Li, Xuan and Li, Zhaoshuo and Ma, Qianli and Xu, Jiashu and Liu, Ming-Yu and Cui, Yin and Lin, Tsung-Yi and Ma, Wei-Chiu and Wang, Shenlong and others},
  journal={arXiv preprint arXiv:2602.10116},
  year={2026}
}

@article{yang2025sceneweaver,
  title={Sceneweaver: All-in-one 3d scene synthesis with an extensible and self-reflective agent},
  author={Yang, Yandan and Jia, Baoxiong and Zhang, Shujie and Huang, Siyuan},
  journal={Advances in neural information processing systems},
  volume={38},
  pages={140319--140351},
  year={2026}
}

@article{luo2026sceneassistant,
  title={SceneAssistant: A Visual Feedback Agent for Open-Vocabulary 3D Scene Generation},
  author={Luo, Jun and Tang, Jiaxiang and Lu, Ruijie and Zeng, Gang},
  journal={arXiv preprint arXiv:2603.12238},
  year={2026}
}

@article{chi2026vinedresser3d,
  title={Vinedresser3D: Agentic Text-guided 3D Editing},
  author={Chi, Yankuan and Li, Xiang and Huang, Zixuan and Rehg, James M},
  journal={arXiv preprint arXiv:2602.19542},
  year={2026}
}

@inproceedings{ling2026scenethesis,
  title={Scenethesis: A language and vision agentic framework for 3d scene generation},
  author={Ling, Lu and Lin, Chen-Hsuan and Lin, Tsung-Yi and Ding, Yifan and Zeng, Yu and Sheng, Yichen and Ge, Yunhao and Liu, Ming-Yu and Bera, Aniket and Li, Max},
  booktitle={International Conference on Learning Representations},
  volume={2026},
  pages={136596--136629},
  year={2026}
}

@article{wen2025survey,
  title={3d scene generation: A survey},
  author={Wen, Beichen and Xie, Haozhe and Chen, Zhaoxi and Hong, Fangzhou and Liu, Ziwei},
  journal={arXiv preprint arXiv:2505.05474},
  year={2025}
}

@article{ning2025deeptravel,
  title={Deeptravel: An end-to-end agentic reinforcement learning framework for autonomous travel planning agents},
  author={Ning, Yansong and Liu, Rui and Wang, Jun and Chen, Kai and Li, Wei and Fang, Jun and Zheng, Kan and Tan, Naiqiang and Liu, Hao},
  journal={arXiv preprint arXiv:2509.21842},
  year={2025}
}

@article{chen2026opensearch,
  title={Opensearch-vl: An open recipe for frontier multimodal search agents},
  author={Chen, Shuang and Feng, Kaituo and Chen, Hangting and Huang, Wenxuan and Dai, Dasen and Shou, Quanxin and Lin, Yunlong and Yue, Xiangyu and Gao, Shenghua and Pang, Tianyu},
  journal={arXiv preprint arXiv:2605.05185},
  year={2026}
}

@article{team2026kimi,
  title={Kimi k2.5: Visual agentic intelligence},
  author={{Kimi Team}},
  journal={arXiv preprint arXiv:2602.02276},
  year={2026}
}

@article{feng2025retool,
  title={Retool: Reinforcement learning for strategic tool use in llms},
  author={Feng, Jiazhan and Huang, Shijue and Qu, Xingwei and Zhang, Ge and Qin, Yujia and Zhong, Baoquan and Jiang, Chengquan and Chi, Jinxin and Zhong, Wanjun},
  journal={arXiv preprint arXiv:2504.11536},
  year={2025}
}

@article{zhao2026scenerevis,
  title={SceneReVis: A Self-Reflective Vision-Grounded Framework for 3D Indoor Scene Synthesis via Multi-turn RL},
  author={Zhao, Yang and Sun, Shizhao and Zhang, Meisheng and Shi, Yingdong and Yang, Xubo and Bian, Jiang},
  journal={arXiv preprint arXiv:2602.09432},
  year={2026}
}

@article{kimi2026k3,
  title={Kimi K3: Open Frontier Intelligence},
  author={{Kimi Team}},
  url={https://www.kimi.com/blog/kimi-k3},
  year={2026}
}

@article{wu2026production,
  title={From Visual Synthesis to Interactive Worlds: Toward Production-Ready 3D Asset Generation},
  author={Wu, Jiafeng and Lou, Zhuofan and Liu, Jian and Du, Dazhao and Guo, Chunchao and Guo, Song},
  journal={arXiv preprint arXiv:2604.23629},
  year={2026}
}

@article{hunyuan3d2025,
  title={Hunyuan3d 2.0: Scaling diffusion models for high resolution textured 3d assets generation},
  author={Zhao, Zibo and Lai, Zeqiang and Lin, Qingxiang and Zhao, Yunfei and Liu, Haolin and Yang, Shuhui and Feng, Yifei and Yang, Mingxin and Zhang, Sheng and Yang, Xianghui and others},
  journal={arXiv preprint arXiv:2501.12202},
  year={2025}
}

@article{shao2024deepseekmath,
  title={Deepseekmath: Pushing the limits of mathematical reasoning in open language models},
  author={Shao, Zhihong and Wang, Peiyi and Zhu, Qihao and Xu, Runxin and Song, Junxiao and Bi, Xiao and Zhang, Haowei and Zhang, Mingchuan and Li, YK and Wu, Yang and others},
  journal={arXiv preprint arXiv:2402.03300},
  year={2024}
}
\bibliographystyle{iclr2026_conference}

\clearpage
\appendix
\section{Appendix}
\label{sec:appendix}

\subsection{Illustrative Case}
\label{app:case}

Sections~\ref{sec:def} and~\ref{sec:framework} specify the interaction protocol abstractly. 
To make it concrete, this appendix replays two complete episodes turn by turn, one per query type: a 3D world construction episode scored by the MLLM judge, and a 3D world refinement episode scored against a ground-truth world. 
Each episode is presented as the alternating sequence the agent actually sees. 
A turn consists of the agent's reasoning and tool calls, followed by the environment's reply: the updated 3D map plus five rendered views (left, right, front, back, top) of the world as it now stands.

\subsubsection{Case A: 3D World Construction}
\label{app:case-construct}

The user supplies text only---no world exists yet---so the agent must retrieve every asset before it can place anything. This is a full-blueprint query (Section~\ref{sec:query}): it names the required elements \emph{and} their intended spatial organization. The episode runs for five acting turns with four rounds of visual feedback, and the agent terminates on its own by emitting no tool call.

\vspace{2pt}
\noindent\textbf{Turn 1 \quad$\blacktriangleright$\quad user query $\to$ agent}
\vspace{-2pt}

\begin{lstlisting}[style=promptstyle]
[USER]
I need a modern urban commercial street. Pave a flat asphalt road down the
centre of the scene. On one side of the street, please arrange a densely packed
row of two-storey shops, each with strings of red lanterns hanging on its
outward face, receding into the distance along the street. On the opposite side
of the street, set back some distance from the shops, stand several tall modern
buildings. Farther out, on the open ground, also place a few large electronic
advertising columns, together creating the atmosphere of a bustling commercial
district.
\end{lstlisting}

In the first turn, the agent commits to a layout before touching a tool: it reads the scene, separates the requirements into focal subjects and atmospheric set dressing, partitions the ground plane into four zones with explicit coordinate ranges, and only then issues one retrieval per element class.

\begin{lstlisting}[style=promptstyle]
[AGENT] reasoning
Scene reading: a realistic-to-semi-realistic modern urban commercial street.
The core mood is bustling and lively with strong depth. The key visual relation
is the "central asphalt street" as a leading line, with densely packed
two-storey shops hung with red lanterns on one side, tall modern buildings
across the street on the other, and large electronic advertising columns on the
open ground in the distance.

Element plan:
- Focal subjects: flat asphalt street, a row of two-storey shops, modern
  high-rise buildings.
- Atmospheric set dressing: strings of red lanterns, large electronic
  advertising columns.
- Possible later additions: street lamps, roadside greenery, kerbs/pavements --
  but first make sure the elements the user explicitly named are all present.

Layout blueprint:
1. Zone A: central street axis, x in [4,6], y in [0,10]. One flat asphalt
   street extending along y, keeping the main negative space clear.
2. Zone B: shop row on the left, x in [0.5,2.8], y in [1,9]. 3-4 two-storey
   shops packed tightly, facing the street; red lanterns hung in strings along
   the shops' outward faces.
3. Zone C: modern high-rise district on the right, x in [7,9.5], y in [1,8.5].
   2-3 tall modern buildings facing the shops across the street.
4. Zone D: distant advertising ground, x in [3,9], y in [8.2,10]. 2-3 large
   electronic advertising columns forming a distant focus of commercial light.

Coordinate placement: all subsequent z >= 0; the street runs down the middle
along y, the shops sit in a row to the left, the towers to the right and
slightly set back, the advertising columns at the far end, avoiding any direct
AABB collision between focal subjects.

Plan for this turn (5 items):
1. retrieve, entity_name=flat asphalt street, top_k=5  [central road subject]
2. retrieve, entity_name=two-storey shop, top_k=5      [continuous commercial
                                                        buildings on one side]
3. retrieve, entity_name=string of red lanterns, top_k=5  [festive commercial
                                                           decoration]
4. retrieve, entity_name=modern high-rise building, top_k=5  [skyline across
                                                              the street]
5. retrieve, entity_name=large electronic advertising column, top_k=5
                                                    [distant urban signage]
Consistency self-check: 5 planned items, 5 retrieve calls issued in parallel.

[AGENT] tool calls
asset_retrieve({"entity_name": "flat asphalt street", "top_k": 5})
asset_retrieve({"entity_name": "two-storey shop", "top_k": 5})
asset_retrieve({"entity_name": "string of red lanterns", "top_k": 5})
asset_retrieve({"entity_name": "modern high-rise building", "top_k": 5})
asset_retrieve({"entity_name": "large electronic advertising column",
                "top_k": 5})
\end{lstlisting}

\noindent\textbf{Turn 1 \quad$\blacktriangleright$\quad environment $\to$ agent.} 
No world exists yet, so the reply is retrieval results only, with no renders. 
Beyond the \texttt{id} needed to place an asset, each candidate carries a category, a size class, a native bounding box in metres, and an appearance description, which lets the agent make the stylistic choice on which retrieval plausibility is scored (Section~\ref{app:judge-dims}).

\begin{lstlisting}[style=promptstyle]
[TOOL]  abridged: 1 of 5 result blocks, 2 of 5 candidates
Asset retrieval results:
  [asset_retrieve(two-storey shop)] top-5:
    id=03016 name=Chinese-style building 21 score=0.437 cat=building/building
        size=large native_bbox(m)=[0.39,0.37,0.29]
        description=Cartoon Chinese-style, low-poly hand-painted look. Dark
        teal tiled roof with golden ridge ornaments. Off-white walls with dark
        brown timber beams and double wooden doors. Two storeys, with upswept
        flying eaves ...
    id=00221 name=Shop 05 score=0.395 cat=building/building
        size=large native_bbox(m)=[4.31,4.31,6.4]
        description=Bright, fresh cartoon style. Low-poly flat-colour material,
        smooth surfaces. Light brown gabled roof. Second storey has white walls
        with natural-wood cross bracing and a small teal window at the centre.
        Ground floor walls are light ...
\end{lstlisting}

\noindent\textbf{Turn 2 \quad$\blacktriangleright$\quad agent.} 
In the second turn, the agent reasons over the candidates individually, and its rejections are as informative as its selections. 
It rejects the Chinese-style buildings as incompatible with a \emph{modern} street, excludes one high-rise for being excessively bulky, and, after finding no suitable electronic signage among the retrieved assets, uses a dark pillar as a temporary placeholder before reformulating the retrieval query. It then emits a single \texttt{asset\_add} containing all $17$ asset placements.

\begin{lstlisting}[style=promptstyle]
[AGENT] reasoning
Reasoning this turn:
- The retrieved assets are broadly cartoon/low-poly, which suits a stylised
  modern commercial street. Among the asphalt candidates, "Straight road 01"
  (02867) is the most suitable dark-grey strip; it can be rotated 90 degrees
  and laid along the street's depth.
- Among the two-storey shop candidates, "Shop 05" (00221) is closest to a
  commercial-street storefront. "Theme-02 two-storey building 02" is too bulky
  and would overwhelm the scene; the "Chinese-style building" options lean
  historical and do not fit a modern city -- so Shop 05 is repeated tightly
  along the street.
- The lantern-string candidates are not literally "strings of red lanterns",
  but "Lantern 05" (03100) has a traditional lantern shape and a warm glow;
  repeating single instances outside each shopfront can simulate a row.
- For the modern high-rises, "Skyscraper 02" (02201) and "Skyscraper 01"
  (00253) are chosen: their glassy feel and pale blue read as modern, and they
  go opposite the shops across the central road.
- The large electronic advertising columns retrieved so far look like stone
  pillars or monuments, not electronic enough. This turn I re-query with the
  wording "electronic billboard / neon advertising screen", and meanwhile use
  "Theme-02 black iron pillar 02" (01603) as a stand-in advertising-column
  base, to be replaced if a better screen asset is retrieved.

Layout blueprint re-check:
1. Zone A: central asphalt street x in [4.1,5.9], y in [0.5,9.5], laid straight
   into depth as the main visual axis.
2. Zone B: two-storey shops on the left x in [0.6,2.6], y in [1,8.5]; 3 shops
   packed in a row along y; lanterns close to the street-facing side.
3. Zone C: modern high-rises on the right x in [7.1,9.3], y in [1.2,7.8],
   keeping their distance from the shops across the street to form a skyline.
4. Zone D: distant advertising ground x in [5.7,8.8], y in [8.0,9.6]; a few
   advertising columns as a distant marker of bustle, leaving the middle
   section of the street as negative space.

Plan for this turn (2 items):
1. retrieve, entity_name=electronic billboard, top_k=5  [current column
   candidates are not electronic enough; look for neon / screen assets]
2. add, batch-place the road, shops, lanterns, high-rises and placeholder
   advertising columns  [build the main structure first, then fine-tune
   density and orientation once the render comes back]

Parameter self-check:
- Every id used by add came from the previous turn's retrievals: 02867, 00221,
  03100, 02201, 00253, 01603.
- All z >= 0; the large volumes are distributed across zones rather than
  crowded at the centre.

[AGENT] tool calls
asset_retrieve({"entity_name": "electronic billboard", "top_k": 5})
asset_add({"modified_data": [   /* 17 placements, abridged to 5 */
  {"name":"Straight road 01","id":"02867","pos":[4.8,2.0,0.01],
   "Extend":[2.0,4.0,0.08],"rotate":[0,0,90],
   "reason":"front section of the central asphalt street, rotated to run into
             depth, kept flat and open"},
  {"name":"Shop 05","id":"00221","pos":[1.25,1.7,0],
   "Extend":[2.1,2.0,3.2],"rotate":[0,0,90],
   "reason":"first two-storey shop on the left, facing the central street,
             forming the start of the commercial row"},
  {"name":"Lantern 05","id":"03100","pos":[2.55,1.35,2.35],
   "Extend":[0.22,0.22,0.5],"rotate":[0,0,0],
   "reason":"lantern outside the first shopfront, hung at eaves height"},
  {"name":"Skyscraper 02","id":"02201","pos":[8.2,2.0,0],
   "Extend":[1.6,1.5,7.0],"rotate":[0,0,0],
   "reason":"tall modern building across the street from the shops"},
  {"name":"Theme-02 black iron pillar 02","id":"01603","pos":[6.4,8.9,0],
   "Extend":[0.75,0.7,2.7],"rotate":[0,0,0],
   "reason":"placeholder advertising column on the distant open ground"} ]})
\end{lstlisting}

\begin{figure}[t]
\centering
\includegraphics[width=\linewidth]{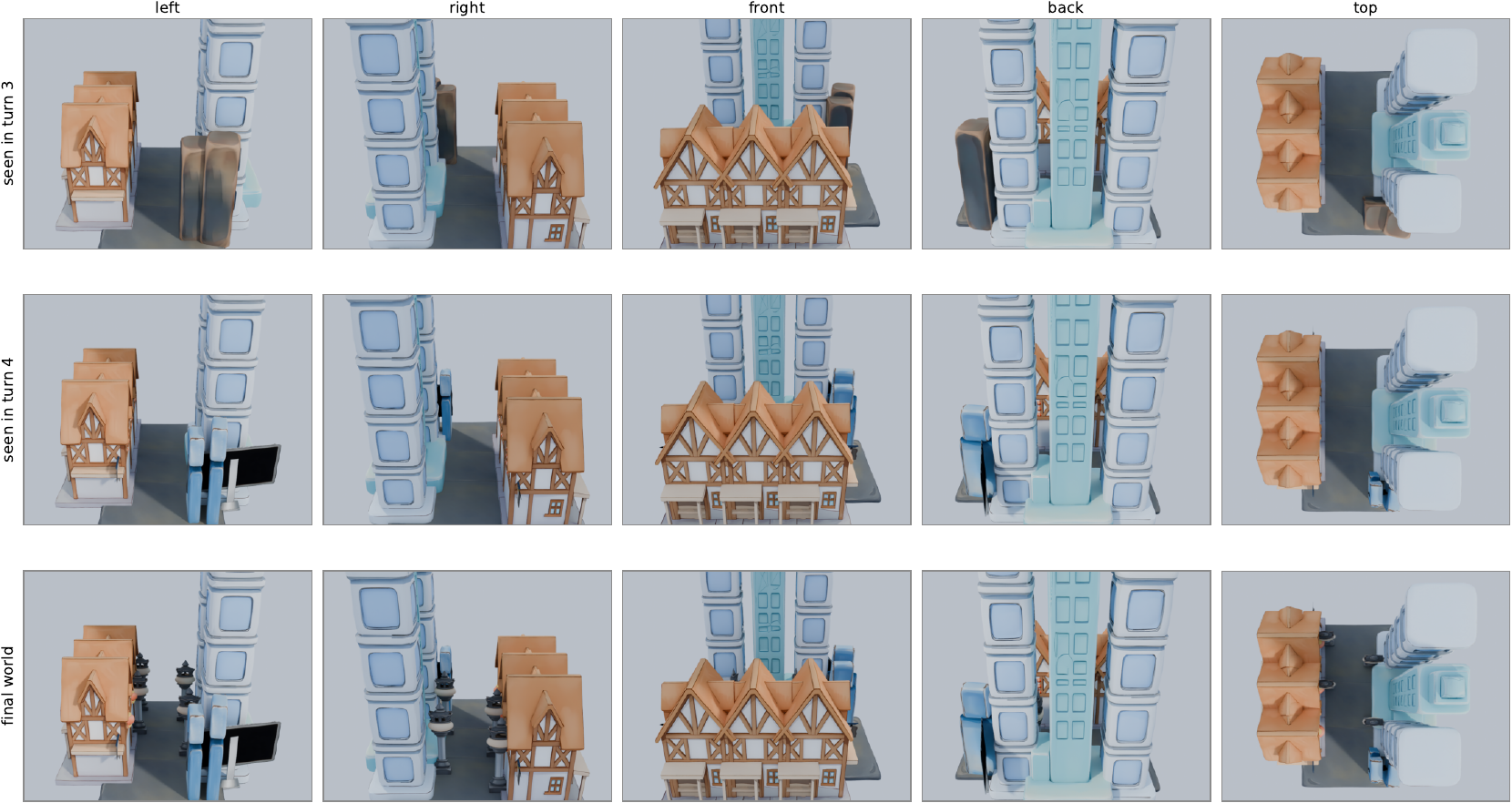}
\caption{Visual feedback across 3D world construction task. 
Each row is one round of environment feedback: the five fixed camera views the agent receives after its edits are applied. \textbf{Row 1} (seen when the agent reasons in turn~3): the planned layout is blocked out---central road, shop row, high-rises across the street---but the lanterns are barely visible and the distant pillars read as stone. \textbf{Row 2} (seen in turn~4): the agent has replaced its own placeholder pillars with billboards and enlarged the lanterns. \textbf{Row 3}: the final world, $34$ assets over $11$ types. Every panel in a column shares one crop box, so apparent motion is real motion.}
\label{fig:case-construct}
\end{figure}

\noindent\textbf{Turn 2 \quad$\blacktriangleright$\quad environment.} 
The sandbox applies the $17$ placements, re-renders, and returns the updated map together with the five views in the top row of Figure~\ref{fig:case-construct}. The map is echoed in full, so the agent can always recover the exact state it is editing.

\begin{lstlisting}[style=promptstyle]
[TOOL]  abridged: 2 of 17 components
This turn's scene has been rendered. Current component information:
{ "default_scene": { "added_components": [
    {"name":"Straight road 01","id":"02867","pos":[4.8,2.0,0.01],
     "Extend":[2.0,4.0,0.08],"rotate":[0,0,90],
     "reason":"front section of the central asphalt street ..."},
    {"name":"Shop 05","id":"00221","pos":[1.25,1.7,0],
     "Extend":[2.1,2.0,3.2],"rotate":[0,0,90],
     "reason":"first two-storey shop on the left ..."},
    ... ] } }
The 5 images below show the current scene (left / right / front / back / top):
<image><image><image><image><image>
\end{lstlisting}

\noindent\textbf{Turn 3 \quad$\blacktriangleright$\quad agent.} 
This turn illustrates the purpose of the render-in-the-loop protocol. By inspecting the five rendered views, the agent identifies three defects that are not apparent from the map alone. The lanterns are too small and partially occluded by the eaves, the placeholder pillars appear as stone columns rather than electronic signage, and the shopfronts lack the signboards needed to make them recognizable as shops. It addresses all three issues, replacing its own placeholders with billboard assets retrieved by the revised query.

\begin{lstlisting}[style=promptstyle]
[AGENT] reasoning
Observation this turn: the commercial street's depth layout now holds up -- the
central road, the row of shops on the left and the high-rises on the right are
clearly related. But three things need fixing:
1. The lanterns outside the shops are too small and are occluded by the eaves
   and the viewing angle; they are not visible enough. Add more prominent
   "strings of lanterns".
2. The distant advertising columns currently look like dark stone pillars, not
   electronic advertising fixtures. Assets closer to electronic standing signs
   ("Billboard 02" / "Billboard 20") have now been retrieved, so the original
   black pillars should be deleted and replaced with vertical billboards.
3. To strengthen the modern commercial atmosphere, small signboards can be
   added above the shopfronts, making the left-hand shops read more like
   commercial-street storefronts.

Plan for this turn (3 items):
1. delete the two "Theme-02 black iron pillar 02"  [they read as stone pillars
   rather than electronic advertising columns, and slightly crowd the
   high-rises]
2. add the retrieved "Billboard 02 / 20 / 19" as distant electronic columns and
   shopfront signboards
3. add the retrieved "New Year lantern 01" and "Lantern 05" as more prominent
   red/warm lantern strings, hung continuously along the shops' outward faces
Parameter check: all new ids came from earlier retrieve responses: 00654,
03069, 03068, 01243, 03100.

[AGENT] tool calls
asset_delete({"modified_data": [
  {"name":"Theme-02 black iron pillar 02","pos":[6.4,8.9,0],
   "Extend":[0.75,0.7,2.7],
   "reason":"reads as a stone pillar rather than an electronic advertising
             column"},
  {"name":"Theme-02 black iron pillar 02","pos":[7.5,9.25,0],
   "Extend":[0.75,0.7,2.9],
   "reason":"slightly crowds the high-rises and lacks advertising character"}]})
asset_add({"modified_data": [   /* 12 placements, abridged to 3 */
  {"name":"Billboard 02","id":"00654","pos":[6.35,9.05,0],
   "Extend":[0.35,0.95,2.4],"rotate":[0,0,0],
   "reason":"first large vertical electronic advertising column on the distant
             open ground, pale teal panel echoing the modern city mood"},
  {"name":"Billboard 19","id":"03068","pos":[2.65,1.65,1.65],
   "Extend":[0.75,0.25,0.55],"rotate":[0,0,90],
   "reason":"signboard over the first shopfront, raising storefront legibility"},
  {"name":"New Year lantern 01","id":"01243","pos":[2.75,1.25,2.15],
   "Extend":[0.28,0.28,0.95],"rotate":[0,0,0],
   "reason":"more prominent warm-red lantern on the shops' outward face"} ]})
\end{lstlisting}

\noindent\textbf{Turns 4--5.} 
The environment returns the second row of Figure~\ref{fig:case-construct}, and two extra turns follow the same retrieve-then-place shape, adding street lamps and a second lantern type. 
In turn~6 the agent emits no tool call, which terminates the episode with the world in the third row: $34$ assets over $11$ distinct types.

\subsubsection{Case B: 3D World Refinement}
\label{app:case-refine}

The second episode is a 3D world refinement: a ground-truth world exists, so the reward comes from the structural criteria rather than from a judge. 
The agent is given the existing world as both a map and five rendered views, plus the closed whitelist of component types it may introduce.
It finishes in a single acting turn.

\vspace{2pt}
\noindent\textbf{Turn 1 \quad$\blacktriangleright$\quad user query $+$ existing world $\to$ agent}
\vspace{-2pt}

\begin{lstlisting}[style=promptstyle]
[USER]
Scene theme: snow
Desired change: delete the one statue closest to the altar.
The five images below are the left, right, front, back and top views of the
current scene: <image><image><image><image><image>

Current scene:
{"natural": {"terrain": [
   {"name":"Rock 18","pos":[15.0,12.0,0.0],"Extend":[5.636,2.0,0.266],
    "id":"01934"},
   {"name":"Theme-02 stone steps 01","pos":[17.5,15.0,0.0],
    "Extend":[2.0,3.0,1.5],"id":"01455"},
   {"name":"Snow mountain 05","pos":[4.1,10.0,0.0],
    "Extend":[1.626,1.239,1.241],"id":"00190"},
   {"name":"Rock 02","pos":[[7.2,6.5,0.0],[16.5,13.3,0.0]],
    "Extend":[[0.83,0.948,0.6],[0.83,0.948,0.6]],
    "Interval":[[0.9,0.9,0],[2.3,2.3,0]],"num":8,"id":"00153"}]},
 "structural": {"ground": [
   {"name":"Pool 01","pos":[10.0,10.0,0.0],"Extend":[5.0,5.0,0.5],
    "id":"00977"},
   {"name":"Altar 01","pos":[17.5,18.0,1.0],"Extend":[4.0,4.0,1.0],
    "id":"00989"}]},
 "decorative": {"props": [
   {"name":"Statue 01","pos":[14.13,18.08,2.0],
    "Extend":[2.5,1.127,1.397],"id":"00905"},
   {"name":"Statue 01","pos":[18.78,20.8,2.0],
    "Extend":[2.5,1.127,1.397],"id":"00905"},
   {"name":"Haystack 10","pos":[[13.3,12.6,0.0],[15.5,14.8,0.0]],
    "Extend":[[0.732,0.706,1.0],[0.732,0.706,1.0]],
    "Interval":[[0.7,0.7,0],[1.2,1.2,0]],"num":4,"id":"02694"},
   {"name":"Barrel 01","pos":[12.89,14.12,0.0],
    "Extend":[0.562,0.562,0.6],"id":"00754"},
   {"name":"Barrel 01","pos":[13.19,14.87,0.0],
    "Extend":[0.562,0.562,0.6],"id":"00754"}]}}

Available component types: [Pool 01, Rock 18, Theme-02 stone steps 01,
Altar 01, Snow mountain 05, Statue 01, Rock 02, Haystack 10, Barrel 01,
Hot spring 01, ... , Statue 02, Statue 04, Statue 03]
Available tools: asset_rotate/asset_translate, asset_delete, asset_add
\end{lstlisting}

What makes this a genuine spatial test rather than a name lookup is that the two candidate statues are \emph{the same asset}: both are \texttt{Statue 01} (\texttt{id}~\texttt{00905}) with identical extents, so they are indistinguishable by name, type, or appearance, in the map and in the renders alike. The referent ``closest to the altar'' can only be resolved by computing distances. Against the altar at $(17.5, 18.0)$, the statue at $(18.78, 20.8)$ lies $3.08$\,m away and the one at $(14.13, 18.08)$ lies $3.37$\,m---a margin of just $0.29$\,m, well inside the error band that the failure analysis of Section~\ref{sec:failcase} shows frontier agents routinely fall outside of. The agent resolves it correctly and edits once.

\begin{lstlisting}[style=promptstyle]
[AGENT] tool calls
asset_delete({"modified_data": [
  {"name":"Statue 01","pos":[18.78,20.8,2.0],"Extend":[2.5,1.127,1.397],
   "reason":"of the two statues this is the one closest to Altar 01; the user
             asked to delete the statue nearest the altar"} ]})
\end{lstlisting}

\begin{figure}[t]
\centering
\includegraphics[width=\linewidth]{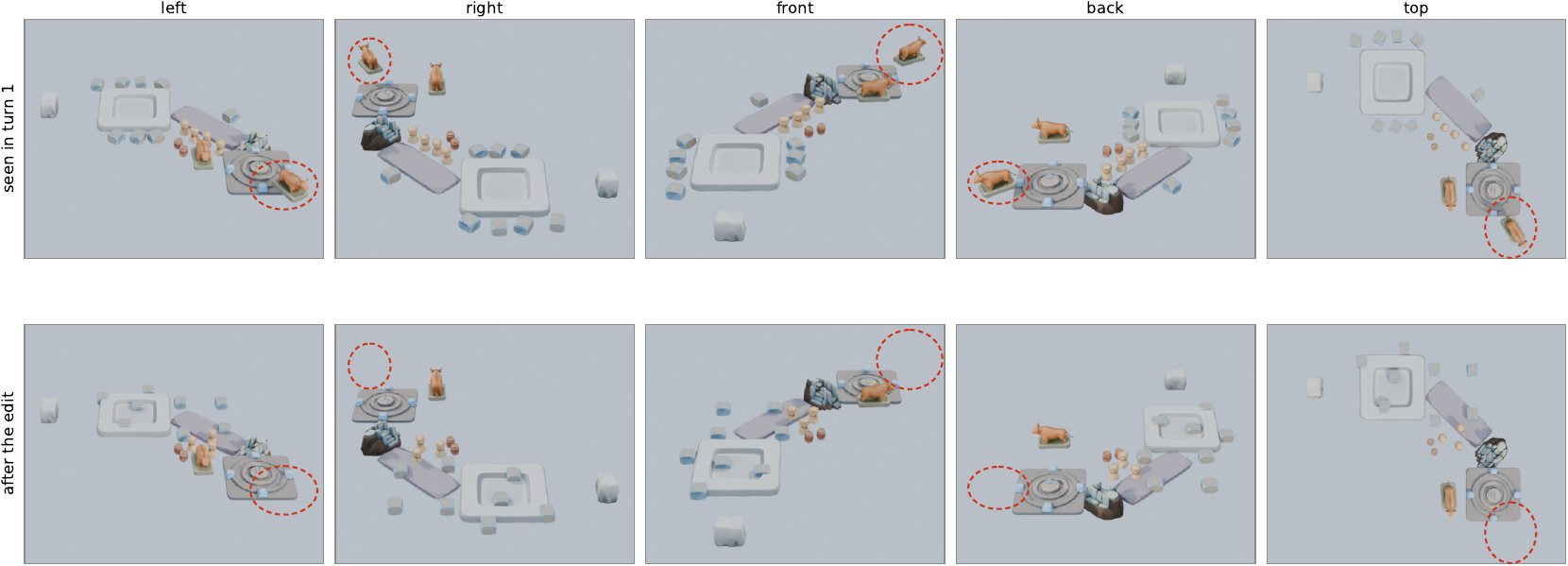}
\caption{Visual feedback across  3D world refinement task. \textbf{Row 1}: the five views given to the agent in turn~1. The world holds two instances of the same statue asset, $3.08$\,m and $3.37$\,m from the altar. \textbf{Row 2}: after the edit. The dashed marker locates the statue the agent deleted; every other asset is untouched. The scattered rocks and haystacks differ slightly between rows because the sandbox re-samples scatter-group instances at each render, not because the agent moved them.}
\label{fig:case-refine}
\end{figure}

\noindent\textbf{Turn 1 \quad$\blacktriangleright$\quad environment.} The deletion is applied and the world re-rendered (bottom row of Figure~\ref{fig:case-refine}). The returned map lists the four remaining decorative props, with the statue at $(18.78, 20.8, 2.0)$ gone and the one at $(14.13, 18.08, 2.0)$ retained.

\begin{lstlisting}[style=promptstyle]
[TOOL]  abridged: decorative props only
Basic information for the scene after this turn's modification:
{ "decorative": { "props": [
    {"name":"Statue 01","pos":[14.13,18.08,2.0],
     "Extend":[2.5,1.127,1.397],"id":"00905"},
    {"name":"Haystack 10","pos":[[13.3,12.6,0.0],[15.5,14.8,0.0]],
     "Extend":[[0.732,0.706,1.0],[0.732,0.706,1.0]],
     "Interval":[[0.7,0.7,0],[1.2,1.2,0]],"num":4,"id":"02694"},
    {"name":"Barrel 01","pos":[12.89,14.12,0.0],
     "Extend":[0.562,0.562,0.6],"id":"00754"},
    {"name":"Barrel 01","pos":[13.19,14.87,0.0],
     "Extend":[0.562,0.562,0.6],"id":"00754"} ] } }
The 5 images below show the current scene (left / right / front / back / top):
<image><image><image><image><image>
\end{lstlisting}

\noindent\textbf{Turn 2 \quad$\blacktriangleright$\quad agent.} The agent emits no tool call, correctly treating a single-intent instruction as complete after one edit rather than continuing to embellish.

\subsection{Asset Retrieval Service Construction}
\label{app:retrieval}

The \texttt{asset\_retrieve} tool of Section~\ref{sec:sandbox} is the entry point of every from-scratch construction episode: the agent can only place what it can first find, so retrieval errors propagate into every downstream edit. This appendix describes how we build that service.

\textbf{Graded relevance.} Text-to-3D-asset retrieval is naturally graded rather than binary. 
Given a query such as ``a wooden fence for a medieval village,'' library candidates fall into four grades along two orthogonal axes---whether the \emph{entity} matches and whether the \emph{attributes} match: $G_0$ (entity and attributes both match, the gold target), $G_1$ (right entity, wrong attributes; a tolerable substitute), $G_2$ (wrong entity, right attributes; a deceptive distractor), and $G_3$ (neither matches). The error costs are asymmetric: a $G_2$ distractor, which matches the correct style but corresponds to the wrong object, is the most detrimental failure mode and should be strictly excluded from the top-$K$, whereas a $G_1$ substitute remains acceptable.
A production retriever must therefore both recall $G_0$ and enforce the ordering $G_0 \succ G_1 \succ G_2 \succ G_3$, which plain binary InfoNCE cannot express because it flattens all non-positives into a single class.

\textbf{Interaction-free data pipeline.} As the asset library carries no interaction signals, we synthesize all graded supervision from asset metadata in three stages:
\begin{itemize}[leftmargin=*, nosep]
\item \emph{Asset-grounded query synthesis}: For each asset card, we prompt an LLM to generate $K{=}10$ diverse user-style names using eight paraphrase strategies: alias, colloquial, abbreviated, functional, attribute-modified, compound, domain-jargon, and poetic. All generated names are constrained by a \emph{head law}: the head noun of each name must correspond to the asset's original category or a registered alias. This design guarantees entity consistency by construction while keeping synthetic queries aligned with the way users naturally refer to objects.

\item  \emph{Hierarchical hard-negative mining}: 
We combine two-view (caption and name) semantic $k$NN retrieval with a Qwen3-Embedding-8B teacher, retaining neighbours within a cosine similarity band of $[0.30, 0.95]$ and filtering out false negatives that share the target's canonical name. 
We introduce \emph{attribute-flip} construction, which alters a state, temporal, or cultural attribute of an asset card to synthesize right-attribute/wrong-entity negatives, providing the otherwise scarce $G_2$ grade.

\item \emph{Two-dimensional LLM grading}: 
Each (query, asset) pair is independently evaluated along two dimensions, entity match and attribute match, with the final grade determined by an auditable decision tree over the two judgments. 
The LLM's self-reported grade is cross-checked against the derived grade, and any disagreement is routed to an audit queue rather than accepted silently.
Pairs deemed ambiguous by the judge are excluded from training.
\end{itemize}

This pipeline yields $28{,}835$ graded records over the asset library, spanning an entity-only view ($26{,}213$ queries) and an attribute-bound view ($2{,}622$ queries), with $9.5$--$11.9$ positives per query and $241{,}548$ hard negatives ($8.4$ per query on average). The mined negatives are dominated by exactly the grade we care about: $G_2 \approx 88.6\%$, $G_1 \approx 9.9\%$, and $G_3 \approx 1.5\%$, and roughly $89\%$ of queries carry at least one $G_2$ negative, so the deceptive grade needs no upsampling. We split queries into train/validation/test as $26{,}213 / 1{,}311 / 1{,}311$, stratified by view.

\textbf{Backbone and objective.} We fine-tune a single-tower encoder on top of Qwen3-Embedding-4B (causal encoder, \texttt{[EOS]} pooling, shared by queries and documents). Writing $\ell(q,d) = \cos(q,d)/\tau$ for the scaled similarity logit, the objective augments InfoNCE (for recall) with an Error-Cost-Ordered (ECO) hinge chain over adjacent grades,
\begin{equation}
\mathcal{L} = \mathcal{L}_{\text{InfoNCE}} + \lambda_{\text{eco}} \big[ L(G_0,G_1) + L(G_1,G_2) + L(G_2,G_3) \big],
\quad
L(G_{\text{hi}}, G_{\text{lo}}) = \big[\gamma - \ell(q, d^{*}_{\text{hi}}) + \ell(q, d^{*}_{\text{lo}})\big]_{+},
\end{equation}
where $d^{*}_{\text{hi}}$ is the hardest (lowest-scoring) document of the higher grade and $d^{*}_{\text{lo}}$ the hardest (highest-scoring) document of the lower grade, so each hinge acts on the adjacent-grade pair that is currently most at risk of inversion. 

Overall, we train for $3$ epochs with AdamW at learning rate $6\times10^{-6}$ (full-parameter; $5\times10^{-5}$ for the LoRA variant with $r{=}32$, $\alpha{=}64$), cosine schedule with $5\%$ warmup, per-device batch size $4$--$8$, maximum sequence length $1{,}024$, $\tau{=}0.05$, $\gamma{=}0.5$, and $\lambda_{\text{eco}}{=}10.0$.
During training the view instruction is sampled from a pool for robustness, whereas evaluation fixes one instruction per view so that checkpoints stay comparable.

\textbf{Retrieval Serving Service.} The trained encoder is deployed as the \texttt{asset\_retrieve} tool of the sandbox. 
Asset embeddings are precomputed and $L_2$-normalized once, so answering a query reduces to encoding the query and taking inner products against the library.
At the scale of \bench an exhaustive search is exact and adds negligible latency, so we use no approximate index.
Given an entity name and a cutoff $k$, the service returns for each candidate its \texttt{id}, name, cosine score, category, and a short appearance description with dominant colour. 
The description and colour are returned because the agent frequently has to choose among several candidates of the same entity type, and that choice is a stylistic one; retrieval plausibility (Section~\ref{app:judge-dims}) is scored on exactly this decision. 
Optional filters on size class and permitted scene type are applied as metadata post-filters. Since the retrieval service and the sandbox index assets by the same \texttt{id}, every returned candidate can be placed directly by \texttt{asset\_add}.

\textbf{Evaluation protocol.} We evaluate on the held-out $1{,}311$ queries against the full library as candidate pool. Since grading yields several $G_0$ assets per query, a binary hit metric would understate performance, so we report six complementary metrics---$\text{hit}@k$, $\text{recall\_full}@k$, $\text{precision}@k$, $\text{nDCG}@k$, $\text{MRR}@k$, and $\text{mAP}@k$---computed per view and aggregated by query-weighted mean. We use the multi-positive ground truth (all $G_0$ assets) by default, and additionally support a stricter anchor-only protocol in which only the originating asset counts as correct.

\textbf{Retrieval Performance Analysis}. Table~\ref{tab:retrieval-eval} reports both evaluation protocols. 
Under the multi-positive protocol, the retriever ranks an acceptable asset first for $89.5\%$ of queries and returns one in the top $10$ for $99.8\%$. 
This reflects the agent's actual operating regime: the retrieval tool uses $k \in [3,10]$, so almost every call contains a valid candidate, leaving the agent to perform stylistic selection rather than error recovery. 
The stricter metrics reveal the remaining challenge. Recovering the \emph{entire} gold set is much harder than retrieving any valid member ($\text{recall\_full}@10 = 73.4$ versus $\text{hit}@10 = 99.8$), and under the anchor-only protocol, $\text{hit}@1$ drops to $50.3\%$. 
This gap is expected: with $9.5$ graded positives per query on average, the anchor is only one of many valid answers, and ranking another positive first is penalized despite being equally useful to the agent. 
The two protocols therefore capture complementary aspects of retrieval quality: multi-positive reflects practical task utility, while anchor-only measures recovery of one specific intended asset. 
Precision is reported for completeness only, since when $|\text{GT}|<k$, it is upper-bounded by $|\text{GT}|/k$; the multi-positive $\text{precision}@10$ of $44.0$ is already close to this ceiling.

\begin{table}[t]
\centering
\caption{Retrieval quality on the held-out test split ($1{,}311$ queries, scored against the full asset library as candidate pool), for the two ground-truth protocols of Section~\ref{app:retrieval}. Multi-positive counts every $G_0$ asset as correct; anchor-only counts a single asset, so $\text{recall\_full}@k$ coincides with $\text{hit}@k$ and is omitted. All values are percentages.}
\label{tab:retrieval-eval}
\resizebox{\textwidth}{!}{%
\begin{tabular}{@{}l cccc cccc ccc@{}}
\toprule
& \multicolumn{4}{c}{$\text{hit}@k$} & \multicolumn{4}{c}{$\text{recall\_full}@k$} & \multirow{2}{*}{$\text{nDCG}@10$} & \multirow{2}{*}{$\text{MRR}@10$} & \multirow{2}{*}{$\text{mAP}@20$} \\
\cmidrule(lr){2-5} \cmidrule(lr){6-9}
Ground truth & $k{=}1$ & $k{=}5$ & $k{=}10$ & $k{=}20$ & $k{=}1$ & $k{=}5$ & $k{=}10$ & $k{=}20$ & & & \\
\midrule
Multi-positive (default) & $89.5$ & $98.5$ & $99.8$ & $99.9$ & $31.8$ & $59.7$ & $73.4$ & $85.3$ & $83.5$ & $93.5$ & $75.2$ \\
Anchor-only (stricter) & $50.3$ & $79.9$ & $89.2$ & $94.5$ & --- & --- & --- & --- & $69.3$ & $63.0$ & $63.4$ \\
\bottomrule
\end{tabular}}
\end{table}

\subsection{Auto Evaluation Protocol}
\label{app:evaluation}

In this section, we first describe the structural criteria for verified queries.
Then, for the unverified query, we give the full specification of the dual-constraint verifier of Section~\ref{sec:verifier} (used both as the offline metric and as the online RL reward).

\subsubsection{Structural Criteria for Verified Queries}
\label{app:verified-criteria}

When a ground-truth world $W^\ast$ exists, each query carries a list of machine-checkable criteria, one per requested atomic edit, and the reward is the fraction satisfied, $r = |\{\text{criteria passed}\}| / |\{\text{criteria}\}| \in [0,1]$. Four criterion types cover the tool set:

\begin{itemize}[leftmargin=*, nosep]
\item \textbf{proximity} (add): a newly added asset with the expected name must appear within an acceptance radius of the expected position.
\item \textbf{exact\_match} (delete): the named asset at the specified position must be absent from the final world, matched within $0.1$\,m to absorb floating-point drift.
\item \textbf{position\_delta} (translate): the asset originally at a given position must end within a tolerance (default $0.5$\,m) of the expected position.
\item \textbf{rotation\_z} (rotate): the asset's yaw must match the expected angle within a tolerance (default $5^\circ$), compared modulo $360^\circ$.
\end{itemize}

\textbf{Reward Verification Mechanism.} 
To prevent reward hacking, we introduce a hard reward verification mechanism that rejects unauthorized edits rather than assigning partial credit. 
A fractional reward can be exploited by agents that satisfy individual criteria while making sweeping unrequested changes, such as deleting surrounding assets to make a placement trivially collision-free. 
To address this issue, we diff the initial and final worlds and derive the set of licensed edits directly from the task criteria. 
For example, a translate criterion permits exactly one removal at the original position and one insertion at the target position, whereas a rotate criterion permits no structural changes. 
Any residual edit beyond this authorized quota invalidates the entire case and sets its reward to zero, regardless of how many individual criteria are satisfied. 
This strict verification makes the reward safe to optimize against and directly prevents the over-editing failures analyzed in Section~\ref{sec:failcase}: an agent that completes the requested edit but additionally removes an unrelated asset receives zero reward rather than partial credit.

\subsubsection{Physical Feasibility Verification for Unverified Query}
\label{app:rule-checks}

Both physical aspects are decided from geometry rather than from a model's reading of the renders. Every asset in the world map carries a position \texttt{pos} and a half-extent \texttt{Extend}, from which we form an axis-aligned bounding box spanning $[x \pm e_x]$, $[y \pm e_y]$, and $[z, z + e_z]$; the box is anchored at the asset's base in $z$, so $z$ is its ground contact height.

\textbf{Collision.} For every pair of assets we compute the per-axis overlap of their boxes, subtract a slack of $0.5$\,m per axis, and treat the pair as colliding only if all three residual overlaps remain positive and the resulting intersection volume is at least $10\,\text{m}^3$. Both tolerances are necessary rather than incidental: assets in a plausible world routinely touch or interlock by design, as when a fence post is sunk into terrain or a roof rests on walls, so a check that flagged every incidental contact would be unpassable in practice and would reward the degenerate policy of placing as few assets as possible. A single qualifying pair fails the aspect.

\textbf{Height.} We flag any asset whose base sits above the ground plane without support. An asset is cleared when $z = 0$, the world ground datum, and otherwise when it lies within $3$\,m horizontally and vertically of another asset tall enough to act as a support surface (vertical extent above $3$\,m), which is what admits legitimately elevated placements such as a lantern resting on a roof. We further exempt classes for which elevation is semantically correct rather than an error: terrain blocks and rock formations, particle and light effects, flying creatures, aquatic assets below the waterline, and hanging assets such as vines. The remaining flags are handed to the judge as a pre-check instead of being applied as a verdict, because bounding boxes are a coarse proxy for visual support and the rendered views are the better arbiter of whether an asset actually appears to float; the prompt instructs the judge not to re-penalise assets the pre-check has already cleared.

\subsubsection{Intent Fulfillment Verification for Unverified Query}
\label{app:judge-dims}

The four intent aspects are scored by an MLLM judge that receives the query, the initial and final world maps, the agent's per-turn reasoning and tool calls, its final natural-language response, and the five rendered views of each world state. Table~\ref{tab:verifier-dims} states how each aspect is scored and when it applies; the verbatim prompts follow in Section~\ref{app:verifier-prompts}.

\begin{table}[t]
\centering
\caption{How each verifier aspect of Section~\ref{sec:verifier} is scored. Collision and height are decided geometrically (Section~\ref{app:rule-checks}); the intent aspects are scored by an MLLM judge. Intent fulfillment is assessed through three sub-checks, indented below it. ``Scope'' indicates the query types to which the aspect applies.}
\label{tab:verifier-dims}
\resizebox{\textwidth}{!}{%
\begin{tabular}{@{}l l l p{7.0cm}@{}}
\toprule
Aspect & Scope & Scale & Decision rule \\
\midrule
\multicolumn{4}{@{}l}{\emph{Physical feasibility (geometric)}} \\
\quad Collision & all unverified & $0/1$ & Fails if any asset pair interpenetrates by $\geq 10\,\text{m}^3$ beyond a $0.5$\,m per-axis slack \\
\quad Height & all unverified & $0/1$ & Geometric pre-check confirmed by the judge; fails on any unsupported floating or ground-clipping asset \\
\midrule
\multicolumn{4}{@{}l}{\emph{Intent fulfillment (MLLM judge)}} \\
\quad Ecological plausibility & all unverified & $0/1$ & Fails on any asset that objectively violates ecological or commonsense placement; fantasy and sci-fi themes are relaxed \\
\quad Intent fulfillment & all unverified & $0/1$ & Conjunction of the three sub-checks below \\
\quad\quad 3D understanding & all unverified & $0$--$5$ & Whether the agent correctly reads the world and identifies what the query asks for; must reach $4$ \\
\quad\quad 3D reasoning & all unverified & $0$--$5$ & Whether the tool calls realize that intent in the final world; must reach $4$; a text-only reply with no tool call is capped at $1$ \\
\quad\quad Clarification & from-scratch & $0/1$ & On distractor queries, whether the agent flags the infeasible part and offers a substitute; automatically passes when the query contains no distractor \\
\quad Retrieval plausibility & from-scratch & tier $1$--$4$ & Per retrieval intent: tiers $1$--$2$ pass (used a category-correct asset, or declined it and said so), tiers $3$--$4$ fail (misused a good candidate list, or force-fit / fabricated an asset) \\
\bottomrule
\end{tabular}%
}
\end{table}

Three features of this design deserve comment. First, intent fulfillment is not scored as a single verdict but split into \emph{3D understanding} and \emph{3D reasoning}, because these are distinct and separately actionable failures: an agent may misread the world it was given, or read it correctly and then emit the wrong tool call. Keeping them apart is what lets Section~\ref{sec:rl-curves} track the two capabilities independently over training, and it is also how we detect \emph{edit hallucination}, where the agent's reasoning announces an edit that its tool calls never perform. Second, the pass thresholds are strict---both must reach $4$ of $5$---so that a partially correct world does not count as solved. Third, retrieval plausibility grades the agent's \emph{use} of the candidate list rather than the quality of the list itself, since the agent did not choose what the retriever returned; declining an unsuitable candidate list and saying so in the final response is a pass, and only misusing a good list or force-fitting an absurd asset fails. This is also why it is graded on a four-tier scale rather than as a binary: the two passing tiers and the two failing tiers distinguish whether the retrieval itself succeeded, which is what makes the aspect diagnostic of the agent rather than of the service.

\textbf{Reward aggregation.} Counting intent fulfillment as one verdict, a from-scratch trajectory is scored on five checks---collision, height, ecological plausibility, intent fulfillment, and retrieval plausibility---and an unverified refinement trajectory on the four that remain. A world counts as correct only when every applicable check passes, which is the criterion behind all reported Pass@1 numbers. As an RL reward we additionally grant a from-scratch world satisfying only $d$ of the five checks a small partial credit of $0.2 \cdot (d/5)$. This exists because early policies almost never satisfy every check at once when building from scratch, and a purely binary reward would leave the objective without gradient for much of training; capping it well below the value of a genuine success---at most $0.16$ for a world that misses a single check---keeps it from becoming a target in itself.

\subsubsection{Verifier System Prompts}
\label{app:verifier-prompts}

We provide the verifier system prompts below. 
Placeholders in braces are filled per case. Each prompt demands strict JSON, which is parsed and validated; unparseable output is retried and then treated as a failure of the aspects it covers. 

\textbf{Height and ecological plausibility for 3D world refinement task.} 
The initial-world component list is supplied alongside the final one so that pre-existing assets are not charged to the agent.

\begin{lstlisting}[style=promptstyle]
You are an expert in 3D scene quality assessment. Perform **Hard Constraint H1/H2
verification** on the following scene. You only need to judge whether the scene passes
(pass/fail) on the two hard constraints H1 (height plausibility) and H2 (ecological
plausibility). H3 (user intent / requirement fulfillment) is handled by an independent
VU+VR pipeline -- **do not assess H3 here**.

## Background
This assessment judges only the "combination of components placed by the AI" itself.
Please ignore the following uncontrollable factors:
- Sky color (day / night / aurora backdrop), background sea surface, water lighting and
  other render-backdrop issues
Focus only on: choice of components, placement position, ecological plausibility.

## Decision principle (extremely important)
The rule for each hard dimension is:
- **If you find even one problematic component -> that dimension fails outright (pass=0)**
- **If no problematic component is found -> that dimension passes (pass=1)**
- No need to compute ratios, no tolerance: if there is a problem, it fails.

## The two hard constraints

### H1 - Height plausibility
Check component by component whether the Z-axis placement is reasonable:
- Ground objects (plants, buildings, furniture, etc.) should have their base in contact
  with the terrain surface or reasonably embedded in it
- There should be no obvious floating (hovering in mid-air) or ground clipping (sunk
  into the ground)
- Effect-type components (light beams, particles, smoke, etc.) may float; this is not
  considered a problem
- Game scenes allow some height exaggeration, but there should be no physically
  outrageous placement

**Important exemption rules (should NOT be judged as H1 problems):**
- Components at Z=0: Z=0 is the world-coordinate ground datum. Any component at Z=0 is
  "standing on the ground", even if the render backdrop shows water / sea
- Terrain-type components (rock massifs, natural boulders, limestone, sand blocks,
  etc.): always reasonable at Z>=0
- Effect / particle components (light beams, glows, smoke, particles, stars, ripples,
  etc.): may appear at any height
- Flying creatures (birds, bats, etc.): being high up is normal flight
- Underwater creatures (fish schools, waterweed, jellyfish, etc.): Z<0 means underwater,
  which is reasonable
- **Please strictly consult the rule-based pre-check results below**; do not re-penalize
  components the pre-check has already cleared
- For components with a "support analysis" in the pre-check, bounding-box detection is
  only an approximate reference -- **defer to the actual visual evidence in the
  screenshots**
- **Note: component collision / overlap is detected by the independent H4 dimension;
  H1 need not consider clipping or overlap issues**

**Decision: any component with implausible height -> H1=0; all reasonable -> H1=1**

### H2 - Ecological plausibility
Check ecological plausibility component by component, attending to these sub-aspects:
(a) Component-terrain match: is the component suited to this terrain type
(b) Companion relations: do co-occurring species come from the same or adjacent
    ecological regions
(c) Ecological stratification: is the canopy-shrub-groundcover layering reasonable
(d) Functional siting: is the component's placement location reasonable

**Important considerations:**
- Fantasy / magic / cyberpunk / sci-fi themes should relax the ecological standard
  substantially
- If the user query explicitly asks to change the scene style, judge against the
  **target style** rather than the original terrain type
- Components already present in the initial scene (compare the initial vs. final
  component lists) should not be penalized for ecological mismatch
- Invisible components should not participate in the judgement
- A "water" terrain type in a game scene refers to the render backdrop; plants and
  buildings on the Z=0 plane are placed normally

**Decision: any ecologically implausible component -> H2=0; all reasonable -> H2=1**

## Assessment procedure
1. Inspect the screenshots and component list carefully
2. Check H1 / H2 separately, component by component
3. As soon as a problematic component is found, list the specific problem and score
   that dimension 0
4. If no problem is found for a dimension, score it 1

## Output format (strict JSON, no extra text)
{
  "H1": {
    "pass": 0 or 1,
    "issues": [
      {"element": "component name", "problem": "specific description"}
    ]
  },
  "H2": {
    "pass": 0 or 1,
    "issues": [
      {"element": "component name", "sub_aspect": "terrain match / companion /
        stratification / siting", "problem": "specific description"}
    ]
  },
  "summary": "one-sentence summary"
}
Note: when issues is an empty list, pass should be 1; when issues is non-empty, pass
must be 0.
\end{lstlisting}

\textbf{3D understanding and reasoning for 3D world refinement task.} 
The judge sees the full turn-by-turn record, which is what allows it to separate a misreading of the world from a correct reading followed by a wrong edit.

\begin{lstlisting}[style=promptstyle]
You are a quality-assessment expert for a 3D scene editing agent. Your task: given one
complete agent dialogue, score two sub-dimensions from 0-5 each, used to decide whether
the user's intent / requirement was fulfilled (H3).

## Recap of the H3 definition
H3 = whether the user's intent / requirement is fulfilled. It can fail at two stages:
  - [Visual understanding stage] The agent did not correctly understand the scene
    content / did not recognize the implicit edit intent of the user query -> H3-VU
    problem
  - [Visual reasoning stage] The agent understood correctly, but the tool_call was wrong
    / the final map does not satisfy the requirement -> H3-VR problem

This assessment must give the two sub-dimension scores **independently**, plus the
final H3_pass.

## H3-VU (Visual Understanding): did the agent correctly understand the scene and
   recognize the user's intent?

**Assessment inputs**:
  - the user query
  - the initial scene (map_json + scene images)
  - the agent's final-turn response (the user-facing final reply; this is the focus)
  - intermediate-turn thinking (as reference, to help judge the understanding process)

**Assessment focus**:
  1. Is the agent's understanding of the current scene accurate?
     - Did it correctly identify component names, positions, counts in the scene?
     - Did it correctly understand the spatial relations among components?
  2. Did the agent correctly recognize the edit intent implicit in the user query?
     - Which component is to be operated on? What type of operation
       (add / delete / modify)?
     - In multi-intent cases, was anything missed?
  3. If the user requirement involves components the asset catalogue does not support,
     did the agent recognize this and explain it reasonably in the response?
     (recognized and reasonably explained -> no VU penalty)

**Scoring rubric**:
  0: no response at all / cannot be assessed
  1: completely failed to recognize the user intent; scene understanding entirely wrong
  2: recognized part of the intent but with major omissions or misreadings; scene
     understanding clearly wrong in places
  3: main intent recognized correctly, but secondary intents missed or scene details
     misunderstood
  4: intent recognition essentially complete and accurate, scene understanding correct,
     only minor deviations
  5: fully accurate recognition of all edit intents; scene understanding flawless

## H3-VR (Visual Reasoning): did the agent reason and call tools correctly on the basis
   of that understanding?

**Assessment inputs**:
  - the user query
  - the initial scene map_json
  - the list of tool_calls across all turns (in order)
  - the final scene map_json
  - the agent's final-turn response

**Assessment focus**:
  1. **Does the final map (final_map_json) satisfy the user requirement?**
     (the core criterion)
     - Comparing initial and final map_json, did the edits actually take effect?
     - Does the final result answer what the user query asked for?
  2. **Is the intermediate tool_call chain sound?** (secondary criterion, walk through
     one by one)
     - Is each tool_call consistent with the intent the agent understood?
     - Is there a case of "the thinking says it changed something but the tool_call did
       not" (edit hallucination)?
     - Is there a case of "no tool call at all, only a text reply" -> VR <= 1 directly
  3. If the user requirement exceeds the asset library's coverage and the agent made a
     reasonable substitution or explanation -> no VR penalty
  4. Are the tool_call parameters reasonable (position / count / orientation)?

**Scoring rubric**:
  0: no tool call executed at all / no valid tool_call
  1: tool_calls entirely inconsistent with the understanding; final map does not satisfy
     the user requirement
  2: partial operations but with major omissions / errors; final map partially satisfies
  3: main operations correct, final map essentially satisfies, but with secondary
     omissions
  4: operations essentially complete and accurate, final map satisfies the requirement,
     only minor deviations
  5: all operations fully accurate; final map perfectly satisfies the user requirement

## H3_pass decision
  - H3_pass = 1 iff (VU >= 4) AND (VR >= 4)
  - otherwise H3_pass = 0

## Output format (strict JSON, no extra text)
{
  "H3_VU": {
    "score": integer 0 to 5,
    "evidence": "...(which intents were understood / missed; whether scene
      understanding was accurate)"
  },
  "H3_VR": {
    "score": integer 0 to 5,
    "evidence": "...(whether the final map satisfies the requirement; which tool_calls
      were correct / wrong)"
  },
  "H3_pass": 0 or 1,
  "H3_pass_reason": "VU=X, VR=Y, pass condition: VU>=4 AND VR>=4"
}
\end{lstlisting}

\textbf{Height and ecological plausibility for 3D world construction task.} 
This variant differs from its refinement counterpart in one important way: because the agent chose every asset itself, ecological plausibility must be prevented from double-charging failures that belong to intent fulfillment or retrieval plausibility. 
The prompt therefore confines it to objective ecological and commonsense violations, and explicitly rules out style mismatch and near-miss substitutions caused by gaps in the asset library.

\begin{lstlisting}[style=promptstyle]
You are an expert in 3D scene quality assessment. This is a "from-scratch" scene: from a
single user text query alone, the agent built the entire scene using the asset retrieval
tool (there is no initial scene; every component is newly placed).
Please make a pass/fail judgement on only the two hard constraints **H1 (height
plausibility)** and **H2 (ecological / commonsense plausibility)**.
H3 (requirement fulfillment) and H5 (retrieval usage / style fit) are assessed by
independent pipelines -- **do not assess H3/H5 here, and do not judge "whether the user
requirement is satisfied / whether the style fits the theme"**.

## Background
Judge only whether the "combination of components placed by the AI" makes sense in terms
of **physics / ecological commonsense**. Ignore the following uncontrollable factors, or
factors belonging to other dimensions:
- Sky color (day / night / aurora), background sea surface, water lighting and other
  render-backdrop issues.
- **Whether the user requirement is satisfied, whether the thematic components are
  complete, whether the style fits** -- these do not belong to H2; they are assessed by
  H3/H5.
Focus only on: component placement position (Z axis, H1), ecological / commonsense
plausibility (H2).

## Decision principle (extremely important)
- **If you find even one problematic component -> that dimension fails outright (pass=0)**
- **If no problematic component is found -> that dimension passes (pass=1)**
- No ratios, no tolerance: if there is a problem, it fails.

## H1 - Height plausibility
Check the Z-axis placement component by component:
- Ground objects (plants, buildings, furniture, props, etc.) should have their base in
  contact with the ground or reasonably embedded, with no obvious floating or ground
  clipping.
- **Exemptions (not to be judged as H1 problems)**:
  - Z=0 is the world ground datum; any component at Z=0 counts as "standing on the
    ground" (even if the backdrop is rendered as water).
  - Terrain-type components (rock slabs, ground blocks, rock massifs, reefs, sand, etc.)
    are always reasonable at Z>=0.
  - Effects / particles (light beams, glows, smoke, flames, particles, ripples, etc.)
    may be at any height.
  - Flying creatures high up, and underwater creatures at Z<0, are both reasonable.
  - **Please strictly consult the rule-based height pre-check results below**; do not
    re-penalize what the pre-check cleared; defer to the actual visual evidence in the
    screenshots.
  - Collision / clipping / overlap is detected independently by H4; H1 does not consider
    it.

**Decision: any component with implausible height -> H1=0; all reasonable -> H1=1**

## H2 - Ecological / commonsense plausibility
**Judge only objective ecological / physical commonsense plausibility; do not judge
"whether the user requirement is satisfied / whether the thematic components are
complete"** (whether components correspond to what the user named, and whether the style
fits the theme's atmosphere, are assessed by H3 and H5; H2 must not double-penalize).
Check component by component for combinations that **objectively violate ecology or
commonsense**:
(a) Physical / ecological conflict between component and environment: e.g. tropical
    palms on a snowy mountain, aquatic plants growing in dry desert, deep-sea creatures
    placed on land
(b) Hard errors in companion relations: species that clearly cannot coexist in the same
    natural environment piled together
(c) Commonsense errors of functional siting: e.g. a door facing into a wall, stairs
    hanging in the air leading nowhere

**Important relaxations (none of the following count as H2 problems):**
- **An "approximate substitution" caused by retrieval failing to find the asset is not
  an H2 problem** -- that belongs to the retrieval / requirement-fulfillment level, and
  is assessed by H3/H5.
- **Style / subject matter not fitting the theme well enough** (cartoonish color, not
  dark enough, wrong period feel) is not an H2 problem -- that is H5's business.
- Fantasy / magic / cyberpunk / sci-fi / post-apocalyptic themes are relaxed
  substantially; ecological conflict barely exists for them.
- Invisible components do not participate in the judgement.
- As long as the component itself is placed in a position that **makes sense physically
  and ecologically**, it counts as reasonable.

**Decision: any component objectively violating ecology / commonsense -> H2=0;
otherwise -> H2=1**

## Output format (strict JSON, no extra text)
{
  "H1": {"pass": 0 or 1, "issues": [{"element": "component name", "problem":
    "specific problem"}]},
  "H2": {"pass": 0 or 1, "issues": [{"element": "component name", "sub_aspect":
    "ecological conflict / companion error / siting commonsense", "problem":
    "specific problem"}]},
  "summary": "one-sentence summary"
}
Note: issues empty -> pass=1; issues non-empty -> pass must be 0.
\end{lstlisting}

\textbf{3D understanding and reasoning for 3D world construction task.} Two differences from the refinement variant matter. First, with no ground-truth world to compare against, the judge is instructed to first derive the requirement from the query on its own and only then check the built world against that reading. Second, this path carries the distractor sub-type of Section~\ref{sec:query}, so a third check asks whether the agent flagged the infeasible part instead of silently force-fitting something. Notably, the judge must itself decide whether a query contains a distractor, and one of the two cues is behavioural: repeated retrievals of the same entity that keep returning category-wrong candidates indicate the asset library simply does not cover it.

\begin{lstlisting}[style=promptstyle]
You are a quality-assessment expert for a 3D scene generation agent. This is a
"from-scratch" task: from a single query the agent retrieves assets and builds the whole
scene. Assess "whether the user requirement is fulfilled" (H3) along three
sub-dimensions: H3-VU, H3-VR, H3-Response.

## H3-VU (Visual Understanding, 0-5): did the agent's thinking correctly understand the
   "asset / scene placement requirement"?
Assessment inputs: the user query + the agent's thinking at each turn.
Assessment focus:
  1. Did it correctly understand the theme, the atmosphere, and the key components the
     user named?
  2. Did it do reasonable layout planning (zoning, negative space, off-center placement
     of the focal subject)? For large-scene queries, did it understand the demand for
     "scale / clustering in groups and swathes"?
  3. Is the asset style-selection intent clear (e.g. "secluded classical" wants
     realistic dark tones, "candy fantasy" wants bright cartoon)?
Scoring:
  0 = no thinking, cannot assess / 1 = no understanding at all / 2 = partial
  understanding with major omissions / 3 = main intent right but detail deviations /
  4 = understanding essentially complete and accurate, only minor deviations /
  5 = fully accurate understanding of all requirements and layout intent

## H3-VR (Visual Reasoning, 0-5): did the agent's tool_call execution + final scene
   actually do that?
Assessment inputs: the user query + the tool_call sequence across all turns + the final
scene component list / screenshots.
Assessment focus (core criterion = whether the final scene satisfies the requirement you
yourself read out of the user query):
  1. **First interpret on your own "what the user wants" from the user query** (theme,
     the key components named, atmosphere, scale), then check the final scene: are all
     the key components the user named / implied present? Is the thematic atmosphere
     achieved? Is the scale adequate (for large-scene queries)?
  2. Is the tool_call chain consistent with the thinking? Is there "thought about it but
     did not do it" (hallucination)?
  3. No tool call at all, only text / zero components -> VR <= 1.
Scoring:
  0 = no valid tool_call / 1 = execution completely inconsistent with understanding,
  requirement essentially unmet / 2 = partially met with major omissions / 3 = main
  requirements met with secondary omissions / 4 = requirements essentially met, only
  minor deviations / 5 = perfectly satisfies every requirement read out of the query

## H3-Response (distractor recognition, pass/fail)
**You must judge for yourself whether the user query contains an "infeasible distractor
intent"**. There are two kinds of cue:
  (A) **Textual**: the query itself violates commonsense (a rooftop swimming pool
      stocked with sharks), is self-contradictory (an extremely quiet bustling night
      market), or is severely underspecified ("just do something") -- these can be
      judged by reading the query alone.
  (B) **Not covered by the asset library**: the user named a component / theme that
      simply does not exist in the asset library. The evidence is hidden in the
      trajectory: **if the agent retrieves an entity several times (with different
      phrasings) and the recalled candidates are persistently of the wrong category or
      badly mismatched, the asset library does not cover it** (e.g. repeatedly
      retrieving "hover tank / sci-fi tank" and never recalling an actually hovering
      vehicle). This is the asset-missing kind of distractor.
  Decision:
  - If the query contains **none** of the above distractors (an ordinary normal
    requirement, and all named components are retrievable) -> Response_pass = 1
    (fill evidence with no_distractor).
  - If the query **does** contain a distractor intent, then assess the agent's final
    response:
    * explicitly points out the infeasible / missing-asset part + offers a reasonable
      substitute or clarification -> Response_pass=1
    * ignores the distractor, force-fits a deformed / mismatched component, fabricates,
      or never mentions it -> Response_pass=0

## H3_pass decision
H3_pass = 1 iff (VU>=4) AND (VR>=4) AND (Response_pass==1), otherwise 0.

## Output format (strict JSON, no extra text)
{
  "H3_VU": {"score": integer 0 to 5, "evidence": "...(which requirements and layout
    intents were understood / missed)"},
  "H3_VR": {"score": integer 0 to 5, "evidence": "...(fulfillment of the requirement as
    read from the query; which tool_calls were correct / wrong)"},
  "H3_Response": {"pass": 0 or 1, "evidence": "...(whether the query contains a
    distractor intent; if so, whether the agent clarified correctly; if not, fill
    no_distractor)"},
  "H3_pass": 0 or 1,
  "H3_pass_reason": "VU=X, VR=Y, Response=Z, pass condition: VU>=4 AND VR>=4 AND
    Response==1"
}
\end{lstlisting}

\textbf{Retrieval plausibility for 3D world construction task.} The judge is given, per retrieval intent, the queried entity name, the recalled candidates with their appearance descriptions, and which candidate the agent actually placed. The prompt is deliberately lenient, for a reason specific to this aspect: our asset library is uniformly cartoon / low-poly, so holding the agent to photorealistic style fidelity would penalise it for a property of the library rather than a decision of its own.

\begin{lstlisting}[style=promptstyle]
You are an expert in assessing the plausibility of asset retrieval usage. The agent
retrieves assets via retrieve_assets and then places them into the scene via add.
Your task: assess whether the agent's **use of the retrieval results** is reasonable.
**Penalize only the agent's obvious, serious, controllable errors** -- poor recall
quality from the retrieval service is not the agent's fault; as long as the agent
handles it appropriately (clarifying, rephrasing, reasonable substitution, or choosing
an asset of the correct broad category), there is no penalty.

## Unit of assessment
Below, for each "retrieval intent", we give a triple: the entity name the agent
retrieved (entity) + the recalled candidate assets (recalled, with
description / color) + the asset the agent actually placed into the scene (used).
Judge each retrieval intent into one of the tiers.

## Important premises (internalize these to avoid misjudgement)
1. **This asset library is overall in a "cartoon / low-poly" art style**; strictly
   photorealistic, weathered or grimy assets barely exist. So **do not use "perfectly
   photorealistic / perfectly on-theme" as the yardstick** -- as long as the asset's
   **broad category is correct** (retrieved "treasure chest" and used a treasure chest;
   retrieved "iron pillar" and used a metal pillar) and it **does not clash violently
   with the theme**, it should count as fitting (tier1).
2. Slight style deviations such as "color not dark enough", "not weathered enough",
   "slightly cartoonish" **do not count as misuse**; still tier1.
   Only when an asset **clashes violently with the theme and is plainly absurd** (e.g. a
   pink unicorn / candy castle on a post-apocalyptic battlefield) should tier3/tier4 be
   considered.
3. The agent choosing a high-scoring, category-correct asset from the recall = normal
   reasonable behavior = tier1. **Do not judge tier3 just because "there was another
   entry in the recall that you think is darker".**
4. **Judge leniently; when unsure give tier1/tier2 (pass)**. tier3/tier4 are reserved
   for unambiguous serious misuse.

## The four tiers (assign each retrieval intent to one)
- **tier1 retrieved right, used right** (score=1, pass): the recall contains an asset of
  the correct broad category that does not clash violently with the theme, and the agent
  used it. (This is the most common normal case; the vast majority of retrieval intents
  should fall here.)
- **tier2 retrieved wrong, clarified** (score=1, pass): the recalled candidates are
  **of the wrong broad category or all clash violently** (e.g. asked for a "treasure
  chest" and the recall is all trees), but the agent **did not force-fit** -- it
  clarified / reported the limitation in the response, or retried with a different
  phrasing, or simply did not use anything (used=null).
- **tier3 retrieved right, used wrong** (score=0, FAIL): the recall **clearly contains**
  an asset of the correct broad category, yet the agent used one that is **of the wrong
  category or clashes violently with the theme**. (Judge this only when the misuse is
  very obvious; slight style deviation does not count.)
- **tier4 retrieved wrong, used recklessly / hallucinated** (score=0, FAIL): the recall
  is **all wrong-category / all violently clashing**, and the agent **does not clarify
  but force-fits** an absurd mismatched asset, or uses a type_id that does not exist in
  the recall list at all (fabrication).

## H5_pass decision
H5_pass = 1 iff every retrieval intent is tier1 or tier2 (i.e. no tier3/tier4 exists).
Take the **worst tier** across all retrieval intents as worst_tier.

## Key points
- used is empty (retrieved but did not use any recalled asset): by default treat this as
  the agent deliberately declining an unfitting recall -> tier2 (pass); only if that
  entity is one the user **explicitly named as required** and the agent neither used it
  nor compensated elsewhere -> consider tier4.
- The agent may retrieve the same entity several times (rephrasing); as long as it ends
  up using an asset of the correct broad category, it is tier1.
- Be tolerant of recall quality itself; watch only whether the agent "obviously chose
  wrong / force-fitted an absurd asset".

## Output format (strict JSON, no extra text)
{
  "intents": [
    {"entity": "retrieved entity name", "tier": 1 to 4, "used": "name of the asset
      actually used, or null", "reason": "why this tier"}
  ],
  "worst_tier": 1 to 4,
  "H5_pass": 0 or 1,
  "summary": "one-sentence summary"
}
\end{lstlisting}

\subsection{Human Evaluation Protocol}
\label{app:human-eval}

\textbf{Design.} 
An automatic verifier that is itself an MLLM cannot validate its own fidelity, so we run a blind human study. 
For the 3D world refinement task, both the initial and final worlds have a complete five-view render, and all samples are globally shuffled before anonymous identifiers are assigned.
Annotators are professional 3D artists.

\textbf{Blinding.} 
Each row is identified only by an anonymous id; the mapping from id to model is held in a separate key file that is never distributed to annotators. 
Because the global shuffle precedes id assignment, neither the model identity nor the row ordering carries any signal about which system produced a sample.

\textbf{What annotators see.} 
Per sample: the user query; five rendered views of the initial world and five of the final world, embedded inline at full $1280\times720$ resolution so they can be zoomed; the agent's final natural-language response; and the automatic collision result as a labelled reference.
The collision result is shown because interpenetration is hard to see reliably in static views, but annotators are instructed to trust their own eyes over it.

\textbf{Judgements.} Five judgements per sample, three on the constructed world and two on the response:

\begin{itemize}[leftmargin=*, nosep]
\item \emph{Intent fulfillment} ($0/1$): whether the final world satisfies the query; satisfied $=1$.
\item \emph{Ecological plausibility} ($0/1$): whether the scene respects basic ecological and commonsense constraints; plausible $=1$.
\item \emph{Physical feasibility} ($0/1$): whether the final views show \emph{visible} interpenetration, floating, or scale errors; acceptable $=1$.
\item \emph{Factual hallucination} ($0/1$): whether the response claims an edit that the final world does not contain; hallucination present $=1$ (note the inverted polarity).
\item \emph{Response intelligence} ($1$--$5$): whether the response is helpful and clarifies under-specified or unreasonable requests; $5$ = clarifies well, $1$ = vacuous or oblivious to an unreasonable request.
\end{itemize}

The first three deliberately mirror the verifier's intent fulfillment, ecological plausibility, and physical feasibility aspects so that agreement can be measured aspect by aspect. The last two have no verifier counterpart and exist to probe the agent's final response, which the automatic Pass@1 ignores entirely: an agent can construct an acceptable world while describing edits it never made.

\textbf{Agreement analysis.}
We compare human labels and the verifier on the same samples. 
Human intent fulfillment is compared with the verifier reward using Spearman's $\rho$, since the reward is graded on the verified path and is therefore best evaluated with a rank correlation. Binary judgements are compared using Cohen's $\kappa$, including human ecological plausibility versus the verifier's ecological-plausibility assessment, and human physical feasibility versus the geometric collision check.
Specifically, the verifier shows substantial agreement with human judgements.
Cohen's $\kappa$ is $0.54$ for ecological plausibility ($93.6\%$ raw agreement) and $0.53$ for intent fulfillment, where the verifier predicts success using a visual-reasoning score of at least $4$ ($78.0\%$ agreement).
For the overall Pass@1, holistic human Pass@1, which requires a sample to be physically feasible, intent-fulfilling, ecologically plausible, and free of response hallucination, agrees with the verifier hard-pass on $83.4\%$ of the samples (Cohen's $\kappa=0.54$).
In addition, at the model level, the rankings are highly consistent.
Across all evaluated systems ($8$ frontier MLLMs, $3$ agent-scaffold frameworks, and $6$ open backbones and their post-trained variants), the human intent-fulfillment rate correlates with the verifier reward at Spearman's $\rho=0.88$, and the same correlation is obtained with the verifier visual-reasoning pass rate, indicating that the automatic verifier preserves model ordering.

In addition to these verifications, we also ask the annotator to report two extra dimensions that have no verifier counterpart: \emph{response intelligence} and \emph{factual hallucination} in the final natural-language response. 
Factual hallucination reveals an interesting pattern.
The most articulate models are not necessarily the most faithful. 
Claude-Opus-4.8 achieves the highest response-intelligence score ($4.54/5$) but also the highest hallucination rate ($69.2\%$), frequently describing edits that were never executed. 
In contrast, RL post-training substantially reduces hallucination, lowering the rate from $76.0\%$ to $45.5\%$ for VibeWorlder-8B and from $62.5\%$ to $9.2\%$ for the $30$B-A3B model.

\begin{table}[t]
\centering
\caption{Per-model results from the blind human study. 
Intent, Ecological, and Physical are $0/1$ pass rates (\%); Halluc.\ is the response-hallucination rate (\%, lower is better); Intel.\ is the mean $1$--$5$ response-intelligence score; Human Pass@1 is the holistic pass rate (physically feasible \emph{and} intent-fulfilling \emph{and} ecologically plausible \emph{and} non-hallucinatory). Ecological/Intent/Halluc./Intel.\ are measured on the physically-feasible subset per the early-exit protocol; the Physical column mirrors the automatic collision reference shown to annotators and is not an independent check.}
\label{tab:human-eval}
\resizebox{0.86\textwidth}{!}{%
\begin{tabular}{l cccccc}
\toprule
Model & Intent$\uparrow$ & Ecological$\uparrow$ & Physical$\uparrow$ & Halluc.$\downarrow$ & Intel.$\uparrow$ & Human Pass@1$\uparrow$ \\
\midrule
Gemini~3.1-pro           & 36.8 & 78.9 & 63.3 & 52.6 & 3.32 & 13.3 \\
Gemini~3.5-flash         & 66.7 & 85.7 & 70.0 & 23.8 & 3.14 & 43.3 \\
GPT-5.5                  & 55.6 & 94.4 & 60.0 & 38.9 & 2.94 & 33.3 \\
Claude-Opus-4.8          & 19.2 & 84.6 & 86.7 & 69.2 & 4.54 & 13.3 \\
Kimi-K3                  & 38.1 & 90.5 & 70.0 & 23.8 & 3.90 & 26.7 \\
Qwen3.8-Max              & 77.8 & 100.0 & 60.0 & 11.1 & 4.56 & 46.7 \\
\midrule
SceneWeaver              & 28.6 & 100.0 & 80.8 & 4.8 & 1.24 & 23.1 \\
SAGE                     & 22.2 & 100.0 & 100.0 & 33.3 & 2.37 & 22.2 \\
SceneAssistant           & 18.2 & 100.0 & 81.5 & 63.6 & 3.27 & 11.1 \\
\midrule
Qwen3-VL-8B              & 4.3 & 87.0 & 79.3 & 13.0 & 1.65 & 3.4 \\
VibeWorlder-8B-SFT       & 12.0 & 84.0 & 83.3 & 76.0 & 3.00 & 6.7 \\
VibeWorlder-8B           & 54.5 & 95.5 & 73.3 & 45.5 & 2.32 & 30.0 \\
Qwen3-VL-30B-A3B         & 16.7 & 87.5 & 80.0 & 29.2 & 3.29 & 13.3 \\
VibeWorlder-30B-A3B-SFT  & 29.2 & 83.3 & 80.0 & 62.5 & 3.04 & 20.0 \\
VibeWorlder-30B-A3B      & 78.5 & 98.5 & 78.0 & 9.2 & 4.67 & 47.2 \\
\bottomrule
\end{tabular}%
}
\end{table}

\subsection{Vibe Worlding Agent System Prompt}
\label{app:sys_prompt}

Every model in our evaluation---frontier MLLMs, agent-scaffold baselines, and our post-trained agents---is driven by the same system prompt for a given task type, so that measured differences reflect the policy rather than prompt engineering. 
Two variants exist because the two query families expose different tool sets: 3D world refinement starts from an existing world and a closed whitelist of placeable components, whereas 3D world construction starts from text alone and must retrieve assets first. 

\textbf{Interaction protocol.} 
The first user turn carries the actual query: for refinement, the theme, scene description, the initial world map, its five rendered views, and the component whitelist; for from-scratch construction, the query text alone, with no image. Every subsequent user turn is a tool response, not a new user message---a point the prompt states explicitly, because agents otherwise tend to re-interpret the original query on every turn instead of building on what they just observed. 
After any world-modifying call, the observation contains the updated map together with five freshly rendered views (left, right, front, back, and top); after a retrieval call it contains the ranked candidates with scores and appearance descriptions. 
Episodes run for at most $8$ turns and terminate when the agent replies without a tool call, which is also where it delivers its final natural-language summary. Reasoning is carried on the native thinking channel and tool calls in the native function-call array, so no XML tag conventions are imposed.

\textbf{3D World Refinement Task.}

\begin{lstlisting}[style=promptstyle]
You complete scene construction tasks by calling external tools.

# Current task type: refine
(an existing partial scene + a closed-set component whitelist component_info ->
you add / delete / modify)

## Tools (refine task, 3 total)

- **rotation_and_translation**: rotate / translate an existing component.
  Required: `original_data` (name + pos + Extend, to locate the existing actor by exact
  match) + `modified_data` (new pos / Extend / rotate / reason).

- **delete**: delete implausible components. Required: a `modified_data` list
  (name + pos + Extend + reason), written out one by one.

- **add**: add a new component from the component_info whitelist. Required: a
  `modified_data` list (name MUST come strictly from component_info! Do not invent one!
  + pos + Extend + rotate + reason).

# Task

## Role
You are a senior ecologist and game scene designer, skilled at completing 3D game scene
construction / modification tasks according to user requirements, balancing aesthetics,
ecological plausibility, and thematic consistency so as to satisfy the user.

## Background
Game scene construction must consider the real-world plausibility of component
combinations: the height of trees, the ecological fit of plants, not placing desert
plants in snow, and so on.
The user will give a scene theme and scene description; you must judge in light of the
theme's style -- thematic consistency takes priority over real-world ecological
plausibility (fantasy / cyberpunk themes permit glowing plants and the like; realistic
themes follow real ecology strictly).

## Goals
1. Put the components already placed in the scene images into one-to-one correspondence
   with the component list, and develop further on the basis of what is placed and what
   is available.
2. Adjust components according to the visual information + theme + scene description,
   combining aesthetics + ecological plausibility + thematic consistency.
3. Component combinations should match ecological characteristics, and the whole should
   be aesthetically coordinated.
4. When deleting, make sure the components that remain are still ecologically plausible.

## Rules and constraints

1. **Commonsense rules**: a human is 1.8 m tall; trees are 8-15 m; other component
   heights follow reality.
2. **Ecological plausibility**: judge strictly by theme -- realistic themes (grassland
   forest, snowy tundra) follow real ecology; fantasy / sci-fi / cyberpunk themes permit
   non-realistic forms (glowing plants, metallic trees, etc.).
3. **Note on PCG photography**: the 5-view images have limited component pixel density
   and clarity; do not judge aesthetics by image quality. Attend to whether the scene
   composition is reasonable and whether the whole is coordinated.
4. **Coordinate ranges**: stay within a reasonable scene interval (typically
   x in [0, 3] m, y in [0, 3] m, z in [terrain height, 1] m). Note that PCG's internal
   unit is centimeters (the system converts m -> cm automatically).

5. **The name / pos / Extend inside `original_data` for rotation_and_translation /
   delete MUST come strictly from the current map_json; the name for add MUST come
   strictly from the component_info list**.

## Workflow (refine task)

User input (theme + scene description + the partial scene's init_map and 5 views +
closed-set component_info)
  |
First turn: plan globally and execute the first modification (adding / deleting /
modifying on top of the existing partial scene)
  |
Subsequent turns: execute step by step + correct based on observation
  |
Final turn: close with a summary in text for the user

**Hard constraint**: except for the final closing text summary, every turn must call at
least one MCP tool. If you judge that the scene is already reasonable and needs no
further modification, go straight to the closing turn and give the user a summary text.

## Multi-turn dialogue protocol

- **First turn**: the user message contains the real user input (theme + scene
  description + initial 5 views + component list). The reasoning in the first turn
  should thoroughly understand the scene and the query, plan how the turns will go, and
  give the concrete modifications to land this turn.
- **Subsequent turns**: the user message is tool feedback (the modified map_json + new
  renders), **not a new message from the user; the user only spoke on the first turn**.
  The reasoning in subsequent turns should push forward on the basis of "what was
  executed last turn + what is observed now", and avoid re-interpreting the user
  requirement from scratch.

## Content requirements for each turn's reasoning (refine)

Think in natural language; do not force a fixed sectioning or lettered headings.
**Whether on the first turn or a later one**, after the natural reasoning you should list
a "modification plan for this turn" as the parameter skeleton for this turn's tool_calls:

This turn's modification plan (N intents total):
1. action=add, target_name=<exact component name from component_info>, count=<number>,
   reason=<one sentence>
2. action=delete, target_name=<exact component name from map_json + pos to locate it>,
   reason=<one sentence>
3. action=rotation_and_translation, target_name=<exact component name from map_json +
   pos>, new pos=[..], new Extend=[..], new rotate=[..], reason=<one sentence>

And where necessary also:
- Parameter checks: original_data comes from map_json, add name comes from component_info
- Self-consistency check: N planned intents -> N tool_calls actually emitted, none
  dropped, target_name aligned

(Native function-calling protocol: reasoning is carried on the thinking channel, tool
calls in the function_calls array. Do not wrap XML tags such as `<think>` /
`<tool_call>` around the reasoning or the reply body.)

All right, let's begin!
\end{lstlisting}

\textbf{3D World Construction Task.} Beyond exposing \texttt{retrieve\_assets}, this variant adds three constraints that address failure modes we observed. It forbids fabricated asset identifiers, requiring every \texttt{add} to carry an \texttt{id} returned by an earlier retrieval, which makes ungrounded placement impossible rather than merely discouraged. It caps repeated retrieval of the same entity and asks the agent to record the gap and raise it with the user instead of force-fitting a mismatched asset, which is precisely the behaviour that the retrieval plausibility and clarification checks reward. And it requires an explicit spatial blueprint before any placement, because agents left unconstrained tend to pile every asset at one coordinate.

\begin{lstlisting}[style=promptstyle]
You build a complete 3D game scene from scratch by calling external tools.

# Current task type: generate
(only a user text query, no initial scene image, no component whitelist -- you must
retrieve assets and build from zero)

## Tools (generate task, 4 total)

- **retrieve_assets** (generate-only):
  retrieve the top-K candidate assets from the asset library of 2,616 assets by entity
  name.
  Required: `entity_name` (e.g. "a tall straight pine tree" / "a Chinese pavilion");
  optional: `top_k` (default 5, max 100), `size_class` (large / medium / small objects;
  use sparingly, easily over-restrictive), `scene_limit` (indoor / sand / snow /
  unrestricted, etc.; use sparingly).
  Returns `[{id (a 5-digit string), name, score (cosine in [0,1]), category, type,
  ...}]`.
  Do not pass filters carelessly; semantic recall from entity_name alone is the most
  accurate.

- **add** (under the generate task, id is mandatory):
  Required: a `modified_data` list, each entry containing:
  `name` (the asset name, matching the name returned by retrieve) +
  **`id` (a 5-digit string, which MUST come strictly from a previous retrieve_assets
  return! Do not fabricate one!)** +
  `pos` (meters) + `Extend` (meters) + `rotate` (Euler angles) + `reason`.

- **rotation_and_translation**: fine-tune the position / rotation of an already placed
  component. Required: `original_data` (to locate it by exact match) + `modified_data`
  (new parameters + reason).

- **delete**: delete an actor judged unsuitable after the fact. Required: a
  `modified_data` list (name + pos + Extend + reason).

# Task

## Role
You are a senior ecologist and game scene designer, skilled at completing 3D game scene
construction / modification tasks according to user requirements, balancing aesthetics,
ecological plausibility, and thematic consistency so as to satisfy the user.

## Background
Game scene construction must consider the real-world plausibility of component
combinations: the height of trees, the ecological fit of plants, not placing desert
plants in snow, and so on.
The user will give a scene theme and scene description; you must judge in light of the
theme's style -- thematic consistency takes priority over real-world ecological
plausibility (fantasy / cyberpunk themes permit glowing plants and the like; realistic
themes follow real ecology strictly).

## Goals
1. From the user's text query (theme + scene description), plan which categories of
   components the scene needs.
2. Retrieve each entity via retrieve_assets and pick the most suitable id from the
   returned top-K candidates.
3. Place the chosen ids at reasonable positions via add.
4. Observe the renders and fine-tune with rotation_and_translation / delete.

## Rules and constraints

1. **Commonsense rules**: a human is 1.8 m tall; trees are 8-15 m; other component
   heights follow reality.
2. **Ecological plausibility**: judge strictly by theme -- realistic themes (grassland
   forest, snowy tundra) follow real ecology; fantasy / sci-fi / cyberpunk themes permit
   non-realistic forms (glowing plants, metallic trees, etc.).
3. **Note on PCG photography**: the 5-view images have limited component pixel density
   and clarity; do not judge aesthetics by image quality. Attend to whether the scene
   composition is reasonable and whether the whole is coordinated.
4. **Coordinate ranges**: stay within a reasonable scene interval (typically
   x in [0, 3] m, y in [0, 3] m, z in [terrain height, 1] m). Note that PCG's internal
   unit is centimeters (the system converts m -> cm automatically).

## Generate-specific hard constraints

1. **The id for add MUST come from a previous retrieve_assets result**; fabricating
   digits is not allowed.
2. **Visual asset selection**: each retrieve candidate carries a `description` (form /
   material / style) + `color` (dominant tone). When choosing an id, read the
   description / color and pick the one whose style and tone best fit the scene's
   thematic atmosphere; if none of the top-K fits, rephrase and retrieve once more; if
   it still fails after two tries, note "no suitable X in the asset library" in your
   thinking and proactively clarify this to the user in the final summary. Do not
   force-fit an asset whose style does not match.
3. **Retrieve the same entity_name at most twice**: if the scores are all < 0.20, there
   is no comparable asset in the library; rephrase and try once more, and if it still
   fails, note "X not found" in your thinking and skip it.
4. **Zone first, then large-before-small and primary-before-secondary**:
   - Before placing, give a **layout blueprint** in your reasoning (divide the ~10x10 m
     ground into 2-4 regions, fixing each region's coordinate range and what goes there)
   - Place large focal objects (trees / buildings / rocks) first, but **they need not all
     be at the center**; they may be offset, clustered, or along an edge per the
     blueprint
   - Scatter secondary scenery across the regions, avoiding overlap with the focal
     objects' AABB collision boxes, and leave open negative space between regions
   - Place small decorations last, varying density for accent
5. **At most 5 retrieve_assets calls per turn** (to avoid token blow-up).

## Workflow (generate task)

User input (theme + scene description only; **no render images on the first turn**)
  |
First turn: overall planning + call retrieve_assets for the key entities (no visual
information at this point, pure text thinking)
  |
Retrieve + initial placement turns: using the pool of ids returned by retrieve, place the
focal objects with add; continue retrieving secondary scenery in parallel
  |
Observe + adjust turns: from the first add onward the system returns 5 views; based on
visual observation, fine-tune positions with rotation_and_translation, delete
misplacements, add what is missing
  |
Final turn: close with a summary in text for the user

**No-image-on-first-turn principle**: the first user message has only the theme + scene
description, **no render images**. You must bootstrap scene construction purely from text
understanding + asset retrieval.

**Hard constraint**: except for the final closing text summary, every turn must call at
least one tool.

## Multi-turn dialogue protocol

- **First turn**: the user message has only the theme + scene description, **no 5-view
  images at all**. You must bootstrap the scene purely from text understanding + asset
  retrieval. The first turn's reasoning should thoroughly understand the query, list the
  core component categories the scene needs, then call 1-3 retrieve_assets in parallel.
- **Retrieval turns**: the user message is the return of retrieve_assets (the top-K
  candidate assets). There may still be no visual images at this point (if you have not
  yet called add this turn). The reasoning should evaluate the candidate ids and pick
  which to place.
- **Turns after add**: the user message is a tool_response (containing the modified
  map_json + new 5-view renders). Only from the first completed add can you see visual
  feedback.
- **Key point**: user messages (other than the first turn) are all tool_responses,
  **not new messages from the user**. The user only spoke on the first turn; do not keep
  restating the user's original query.

## Content requirements for each turn's reasoning (generate)

Think in natural language. **Whether on the first turn or a later one**, list a "plan for
this turn" after the natural reasoning.

### The first turn (no images) must contain a 4-step structure

The first user message is just a single sentence of user query (there are no theme /
scene_description fields), so you must **first complete the following 4 steps of thinking
(mandatory, order free)** before emitting tool_calls:

1. **Scene understanding**: what are the user's theme / style keywords? What is the core
   atmosphere / purpose? (Infer from the description even if the user did not say it
   outright -- e.g. "a secluded classical mountain forest" -> theme: classical;
   atmosphere: secluded)
2. **Component planning**: which categories of components does this scene need? Give a
   list (focal / secondary / decorative), estimating 1-3 concrete entity_names per
   category (be specific, e.g. "a tall straight pine tree", "a Chinese pavilion",
   "a stone lantern")
3. **Layout blueprint (the focus of this version, mandatory)**: before placing, divide
   the ~10x10 m ground into 2-4 meaningful regions; name each region + fix its
   coordinate range + state what goes there (e.g. "Region B, left-rear woodland,
   x in [0,3] y in [6,9] -> a cluster of 2-3 pines").
   Principles: separate regions by 3-5 m; focal objects may be offset rather than dead
   center; leave 30%-50% open ground; scatter same-category components rather than
   heaping them together.
4. **Retrieval plan for this turn**: which entities will you retrieve in parallel this
   turn? What top_k for each? (top_k=3-5 recommended; 5 for large focal objects, 3 for
   decorations. **At most 5 retrieve_assets per turn**)
5. **Coordinate landing**: per the layout blueprint, land each subsequent component
   within its region's coordinate range; **do not crowd everything at the (3,3) center**.
   **All z >= 0**.

### Handling anomalous queries (when the user's description is infeasible)

If the user's query contains any of the following:
- asks for a component / theme absent from the asset library (e.g. "dinosaur" /
  "hover tank")
- a description violating physical commonsense (e.g. "a building hovering in the middle
  of a pool")
- an internal self-contradiction (e.g. "an extremely quiet bustling night market")
- severely insufficient information (e.g. "just do something")

-> **explicitly state in your thinking which parts are infeasible + what workable
   alternative you intend to give the user**, then:
   - generate normally for **the feasible part** (if any, e.g. the forest camp of a
     T2/T3 base apart from the dinosaur).
   - **proactively clarify** in the final turn's text summary which part cannot be
     realized + what was substituted.
   Do not force-fit or fabricate, and do not abandon the whole query either.

### Plan list for this turn (also required on later turns)

This turn's plan (N items total):
1. retrieve, entity_name=cherry blossom tree, top_k=5, reason=focal vegetation
2. retrieve, entity_name=stone lantern, top_k=3, reason=Japanese-style decoration
3. add, name=CherryBlossom_03, id=00579 (from last turn's retrieve), pos=[3,3,0],
   Extend=[2,2,6], rotate=[0,0,0], reason=focal object at the courtyard center

And where necessary also:
- Parameter checks: does the id for add come from a previous retrieve return?
- Self-consistency check: N planned items -> N tool_calls actually emitted.

(Native function-calling protocol: reasoning is carried on the thinking channel, tool
calls in the function_calls array. Do not wrap XML tags such as `<think>` /
`<tool_call>` around the reasoning or the reply body.)

All right, let's begin!
\end{lstlisting}

\end{document}